\documentclass{article}

\PassOptionsToPackage{numbers, compress}{natbib}

\usepackage{wrapfig}
\usepackage{multirow}

\usepackage[preprint]{neurips_2026}

\usepackage[utf8]{inputenc} 
\usepackage[T1]{fontenc}    
\usepackage{hyperref}       
\usepackage{url}            
\usepackage{booktabs}       
\usepackage{amsfonts}       
\usepackage{nicefrac}       
\usepackage{microtype}      
\usepackage{xcolor}         
\usepackage{graphicx}
\usepackage[most]{tcolorbox}
\usepackage{enumitem}

\usepackage{caption}
\usepackage[T1]{fontenc} 
\tcbuselibrary{skins, breakable}
\usepackage{alltt}
\definecolor{promptbg}{HTML}{F2F2F2}        
\definecolor{prompttitlebg}{HTML}{2B2B2B}   
\definecolor{promptborder}{HTML}{2B2B2B}    

\newtcolorbox{promptbox}[1][]{
    enhanced,
    breakable,
    colback=promptbg,
    colframe=promptborder,
    boxrule=0.4pt,
    arc=2pt,
    outer arc=2pt,
    left=8pt, right=8pt, top=6pt, bottom=6pt,
    fonttitle=\bfseries\small,
    coltitle=white,
    colbacktitle=prompttitlebg,
    title={#1},
    before skip=8pt, after skip=8pt,
}

\title{Memory-R2: Fair Credit Assignment for Long-Horizon Memory-Augmented LLM Agents}

\author{
 \textbf{Sikuan Yan\textsuperscript{*1,2,3}},
 \textbf{Ahmed Bahloul\textsuperscript{*4}},
 \textbf{Ercong Nie\textsuperscript{1}},
 \textbf{Susanna Schwarzmann\textsuperscript{3}},
\\
 \textbf{Riccardo Trivisonno\textsuperscript{3}},
 \textbf{Volker Tresp\textsuperscript{1,2}},
 \textbf{Yunpu Ma\textsuperscript{\dag}\textsuperscript{1,2}}
\\
 \textsuperscript{1}Ludwig Maximilian University of Munich,
 \textsuperscript{2}Munich Center for Machine Learning,
\\
 \textsuperscript{3}Huawei Heisenberg Research Center (Munich),
 \textsuperscript{4}Technical University of Munich
\\
 \small{
    \href{mailto:email@domain}{s.yan@campus.lmu.de}, \href{mailto:email@domain}{cognitive.yunpu@gmail.com}
 }
}

\begin{document}

\maketitle
\renewcommand{\thefootnote}{\fnsymbol{footnote}}
\footnotetext[1]{Equal contribution.}\footnotetext[2]{Corresponding author. The code is available for access via \href{https://github.com/ahmedehabb/Memory-R2}{this repository}.}

\begin{abstract}
Memory-augmented LLM agents enable interactions that extend beyond finite context windows by storing, updating, and reusing information across sessions. However, training such agents with reinforcement learning in multi-session environments is challenging because memory turns the agent's past actions into part of its future environment. Once different rollouts write, update, or delete different memories, they no longer share the same intermediate memory state, making trajectory-level comparisons fundamentally unfair. This violates a key assumption behind group-relative methods such as GRPO, where rollouts are compared as if they were sampled from the same effective environment. Consequently, trajectory-level rewards provide noisy or biased credit signals for long-horizon memory operations. To address this challenge, we introduce \textbf{Memory-R2}, a training framework for long-horizon memory-augmented LLM agents. Its core algorithm, \textbf{LoGo-GRPO}, combines \underline{lo}cal and \underline{g}l\underline{o}bal group-relative optimization. The global objective preserves end-to-end learning from long-horizon trajectory-level rewards, while local rerollouts compare different memory-operation outcomes from the same intermediate memory state, yielding fairer group comparisons and more precise supervision for memory construction. Beyond credit assignment, Memory-R2 jointly optimizes memory formation and memory evolution with a shared-parameter co-learning design, where a fact extractor and a memory manager are instantiated from the same LLM backbone through role-specific prompts. To stabilize multi-step RL over long memory horizons, we adopt a progressive curriculum that increases the training horizon from 8 to 16 to 32 sessions. Together, these components provide an effective training paradigm for memory-augmented LLM agents in long-horizon multi-session settings.
\end{abstract}

\begin{figure}[t]
  \centering
  \includegraphics[width=1\textwidth]{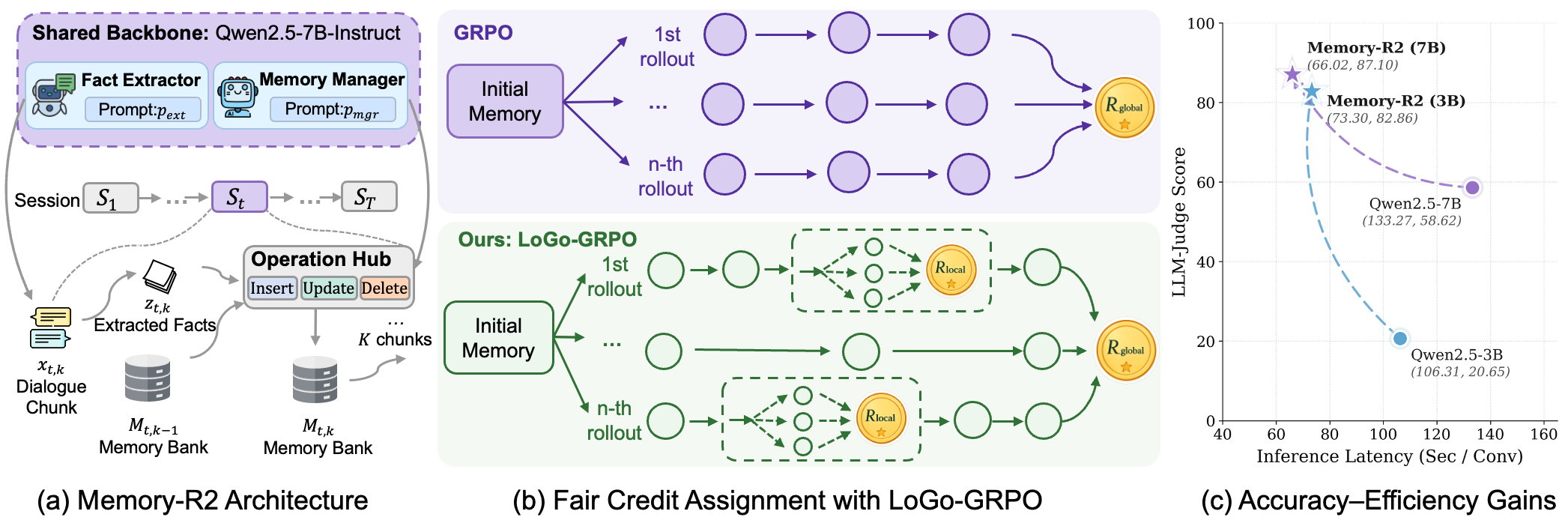}
  \caption{Overview of Memory-R2. (a) Memory-R2 uses a shared-backbone extractor–manager architecture for chunk-wise memory construction. (b) LoGo-GRPO contrasts with standard GRPO by introducing local rerollouts from shared intermediate memory states for fairer credit assignment while preserving global trajectory-level optimization. (c) Memory-R2 improves accuracy and inference latency across backbones.}
  \label{fig:teaser}
\end{figure}

\section{Introduction}
\label{intro}

Large language models (LLMs) have rapidly evolved from standalone text generators into agentic systems that can plan \cite{react-yao2023reactsynergizingreasoningacting}, use tools \cite{toolrl-qian2025toolrlrewardtoollearning, searchr1-jin2025searchr1trainingllmsreason}, and interact over long horizons \cite{webagentr1-wei2025webagentr1trainingwebagents}. A central requirement for such agents is the ability to accumulate, update, and reuse information across interactions. However, despite strong in-context reasoning ability, LLM agents remain fundamentally constrained by finite context windows and the lack of persistent state, making it difficult to retain salient user information, track long-term goals, or maintain consistency over extended multi-session interactions \cite{li2024longcontextllmsstrugglelong, liu2025comprehensivesurveylongcontext}.

To address this limitation, a growing body of work augments LLM agents with explicit memory systems \cite{memorybank-zhong2023memorybankenhancinglargelanguage, amem-xu2025amemagenticmemoryllm}. Existing research broadly follows two directions. The first focuses on memory infrastructure, including graph-structured memory, structured memory schemas, and system-inspired memory organization \cite{zep-rasmussen2025zeptemporalknowledgegraph, mem0-chhikara2025mem0, cam-li2025camconstructivistviewagentic, zhang2025gmemorytracinghierarchicalmemory, memos-li2025memosoperatingmemoryaugmentedgeneration, memoryos-kang2025memoryosaiagent}. The second focuses on memory policy learning, where reinforcement learning (RL) is used to decide what to extract, how to update memory, and how to use retrieved memory \cite{memoryr1-yan2025memory, memalpha-wang2025mem}. While these efforts have substantially improved long-horizon agent behavior, training memory agents in multi-session environments remains fundamentally challenging.

The core difficulty is that memory makes the environment non-stationary. In multi-session agent training, memory turns the agent's past actions into part of its future environment: what the agent writes, updates, or deletes in one session becomes the state inherited by subsequent sessions. This creates a fundamental challenge for trajectory-level RL, especially for group-relative methods such as GRPO~\cite{deepseekr1-Guo_2025}, which rely on comparing rollouts sampled from the same effective environment. Once rollouts modify memory differently, they no longer share the same intermediate memory state, yet GRPO still normalizes their rewards within a single comparison group, leading to unfair comparisons and biased credit assignment. The problem is further amplified by trajectory-level rewards: when a downstream failure occurs, it is difficult to determine whether it comes from the current session's memory operation, corrupted memory inherited from earlier sessions, or later updates that overwrite useful information. This raises a simple but important question:

\begin{tcolorbox}[
    colback=gray!15,
    colframe=black,
    boxrule=0.8pt,
    arc=2mm,
    left=2mm,
    right=2mm,
    top=1mm,
    bottom=1mm
]
\itshape
How can we design a training paradigm for memory-augmented agents that provides more accurate and fair credit assignment across sessions?
\end{tcolorbox}

In this work, we present \textbf{Memory-R2}, a training framework for long-horizon memory-augmented LLM agents, as illustrated in Figure~\ref{fig:teaser}. At its core is \textbf{LoGo-GRPO}, a credit-assignment algorithm that combines \emph{global} and \emph{local} group-relative optimization. LoGo-GRPO preserves a trajectory-level global reward for end-to-end long-horizon optimization, while additionally introducing session-wise attribution signals and local rerollouts that compare trajectories starting from identical intermediate memory states. This yields fairer group comparison and cleaner supervision for memory operations.


Beyond fair credit assignment, Memory-R2 is designed to optimize the whole memory lifecycle. Recent analyses decompose agentic memory into memory formation, memory evolution, and memory retrieval \cite{memsurvey-hu2025memory}, whereas prior RL-based memory work has focused primarily on evolution and retrieval \cite{memoryr1-yan2025memory}. Our framework targets memory formation and evolution through two cooperative roles: a \emph{fact extractor}, which identifies salient information from the interaction context, and a \emph{memory manager}, which decides whether to insert, update, or delete memory entries. Inspired by shared-policy multi-agent RL \cite{rema-wan2025remalearningmetathinkllms}, we instantiate both roles with a shared LLM backbone and role-specific prompts, enabling parameter-efficient co-learning and tighter coordination between extraction and memory editing.

We further formulate memory construction as a multi-step decision process within each session. Rather than treating a session as a single monolithic transition, we divide it into chunks and allow the fact extractor and memory manager to alternate over them, turning memory construction into a temporally extended process that can be refined as more evidence becomes available. To stabilize long-horizon optimization, we also introduce a curriculum over session horizon, progressively scaling training from 8 to 16 to 32 sessions so that the model first acquires reliable short-horizon memory behavior before adapting to more challenging long-context settings.
Our contributions are summarized as follows:
\begin{itemize}[leftmargin=1.1em,labelsep=0.45em,itemsep=0pt,topsep=0pt,parsep=0pt,partopsep=0pt]
\item We propose \textbf{Memory-R2}, a training framework for long-horizon memory-augmented LLM agents, whose core algorithm \textbf{LoGo-GRPO} improves fairness and session-level credit assignment through global-local group-relative optimization.
\item We introduce a \textbf{shared-parameter extractor--manager architecture} and formulate memory construction as a \textbf{multi-step decision process} over chunked sessions, enabling joint optimization of memory formation and evolution.
\item We develop a \textbf{curriculum learning strategy} over session horizon that stabilizes long-horizon RL training, and show that the resulting system is highly \textbf{data-efficient}, achieving strong gains over prior memory-agent baselines using only two training conversations while generalizing across benchmarks, model scales, and answer agents.
\end{itemize}

\section{Related Work}
\label{related_work}

\subsection{Memory Agent Architectures}
Explicit memory has become a standard way to extend LLM agents beyond finite context windows and support long-horizon interaction \cite{amem-xu2025amemagenticmemoryllm,memorybank-zhong2023memorybankenhancinglargelanguage}. Prior work mainly differs in how memory is represented and managed. Representative examples include graph- or structure-based memory systems such as Zep~\cite{zep-rasmussen2025zeptemporalknowledgegraph}, G-Memory~\cite{zhang2025gmemorytracinghierarchicalmemory}, A-MEM~\cite{amem-xu2025amemagenticmemoryllm}, Mem0~\cite{mem0-chhikara2025mem0}, and CAM~\cite{cam-li2025camconstructivistviewagentic}, as well as system-inspired designs such as MemOS \cite{memos-li2025memosoperatingmemoryaugmentedgeneration} and MemoryOS \cite{memoryos-kang2025memoryosaiagent}. While these methods propose increasingly expressive memory substrates, they mostly rely on heuristic or prompt-based policies for deciding what to store, update, or discard. In contrast, our work retains a modular extractor--manager architecture but optimizes the memory lifecycle directly with reinforcement learning.

\subsection{Reinforcement Learning for Memory Agents}
Reinforcement learning has recently become an effective paradigm for training LLM agents in interactive settings such as tool use, web navigation, and reasoning \cite{toolrl-qian2025toolrlrewardtoollearning,searchr1-jin2025searchr1trainingllmsreason,webagentr1-wei2025webagentr1trainingwebagents,deepseekr1-Guo_2025}. This is particularly suitable for memory agents, where the quality of extraction, memory editing, and retrieval decisions is only revealed through downstream task performance. Existing RL-based memory methods, such as Memory-R1~\cite{memoryr1-yan2025memory} and Mem-$\alpha$~\cite{memalpha-wang2025mem}, demonstrate the promise of this direction. However, they rely mainly on outcome-level rewards and do not explicitly address cross-session credit assignment under diverging memory states. They also focus primarily on memory evolution and retrieval, leaving joint optimization of formation, evolution, and retrieval underexplored \cite{memsurvey-hu2025memory}. Our work addresses these gaps by introducing multi-step extractor--manager training, shared-parameter co-learning, and a global-local GRPO objective for fairer credit assignment in long-horizon multi-session settings.

\section{Method}
\label{method}

\subsection{Problem Formulation: Multi-step Memory Bank Construction}
\label{sec:problem_formulation}

We study memory bank construction for long-horizon multi-session interactions. Let $\mathcal{D}=\{S_t\}_{t=1}^{T}$ denote a dialogue trajectory of $T$ sessions, where each session $S_t=\{x_{t,k}\}_{k=1}^{K}$ is divided into $K$ chunks. The agent maintains an external memory bank $\mathcal{M}$ that evolves across sessions. We formulate memory construction as a chunk-wise multi-step process, illustrated in Figure~\ref{fig:teaser}(a): for each chunk $x_{t,k}$, a fact extractor first proposes salient content
\begin{equation}
z_{t,k} \sim \pi_{\mathrm{ext}}\!\left(z \mid x_{t,k}\right),
\label{eq:extractor}
\end{equation}
and a memory manager then chooses an operation conditioned on the extracted content and current memory state,
\begin{equation}
a_{t,k} \sim \pi_{\mathrm{mgr}}\!\left(a \mid z_{t,k}, \mathcal{M}_{t,k-1}\right),
\label{eq:manager}
\end{equation}
where $a_{t,k} \in \mathcal{A}$ denotes operations such as \texttt{INSERT}, \texttt{UPDATE}, and \texttt{DELETE}. The memory bank is updated by a deterministic transition operator
\begin{equation}
\mathcal{M}_{t,k} = \mathcal{T}\!\left(\mathcal{M}_{t,k-1}, z_{t,k}, a_{t,k}\right).
\label{eq:transition}
\end{equation}

This yields a chunk-wise memory construction process over session $t$:
\begin{equation}
\mathcal{M}_{t,0}
\xrightarrow[\pi_{\mathrm{ext}},\,\pi_{\mathrm{mgr}}]{x_{t,1}}
\mathcal{M}_{t,1}
\xrightarrow[\pi_{\mathrm{ext}},\,\pi_{\mathrm{mgr}}]{x_{t,2}}
\cdots
\xrightarrow[\pi_{\mathrm{ext}},\,\pi_{\mathrm{mgr}}]{x_{t,K}}
\mathcal{M}_{t,K},
\label{eq:chunkwise_process}
\end{equation}

Across the full dialogue trajectory, let $\tau=\{z_{t,k},a_{t,k}\}_{t=1,k=1}^{T,K}$ denote a memory-construction rollout. Its probability factorizes as
\begin{equation}
p_{\theta}(\tau \mid \mathcal{D})
=
\prod_{t=1}^{T}\prod_{k=1}^{K}
\pi_{\mathrm{ext}}\!\left(z_{t,k}\mid x_{t,k}\right)
\pi_{\mathrm{mgr}}\!\left(a_{t,k}\mid z_{t,k}, \mathcal{M}_{t,k-1}\right).
\label{eq:trajectory_factorization}
\end{equation}

In our framework, the extractor and the manager are implemented as two cooperative roles instantiated from a shared LLM backbone with role-specific prompts:
\begin{equation}
\pi_{\mathrm{ext}}(\cdot) = \pi_{\theta}(\cdot \mid p_{\mathrm{ext}}, \cdot),
\qquad
\pi_{\mathrm{mgr}}(\cdot) = \pi_{\theta}(\cdot \mid p_{\mathrm{mgr}}, \cdot),
\label{eq:shared_parameter}
\end{equation}
where $\theta$ denotes the shared model parameters, and $p_{\mathrm{ext}}$ and $p_{\mathrm{mgr}}$ are role-specific prompts for fact extraction and memory management, respectively. The resulting memory-construction rollout $\tau$ is evaluated through downstream task performance, yielding a trajectory-level reward $R(\tau)$. We optimize the shared memory policy by maximizing the expected return $\mathbb{E}_{\tau \sim \pi_{\theta}}[R(\tau)]$.

\subsection{Length-Normalized Step-level RL with Shared Extractor--Manager Policy}
\label{sec:multistep_rl}

While Sec.~\ref{sec:problem_formulation} defines memory construction as a multi-step process, optimizing it with a shared LLM policy introduces length-induced bias. We instantiate fact extraction and memory management as two roles of a shared policy with role-specific prompts~\cite{rema-wan2025remalearningmetathinkllms}. Since the two roles produce outputs of different lengths, token-level RL assigns more loss terms to longer generations, biasing the shared policy toward verbose outputs and roles with longer outputs. To address this, we use a \emph{length-normalized step-level} objective, treating each extractor or manager call as one generation step. For a generation step $u$ with generated token indices $\mathcal{U}_u$, we aggregate token-level ratios and advantages as
\begin{equation}
\rho_u
=
\exp\!\left(
\frac{1}{|\mathcal{U}_u|}
\sum_{\ell \in \mathcal{U}_u}
\log
\frac{\pi_{\theta}(y_{\ell}\mid h_{\ell})}
{\pi_{\theta_{\mathrm{old}}}(y_{\ell}\mid h_{\ell})}
\right),
\qquad
\bar{A}_u
=
\frac{1}{|\mathcal{U}_u|}
\sum_{\ell \in \mathcal{U}_u}
A_{\ell},
\label{eq:step_level}
\end{equation}
where $\rho_u$ is the step-level importance ratio, $\bar{A}_u$ is the step-level advantage, $y_{\ell}$ is a generated token, $h_{\ell}$ is its autoregressive context, $A_{\ell}$ is the token-level advantage, and $\pi_{\theta_{\mathrm{old}}}$ is the rollout policy. This gives each generation step comparable weight regardless of output length. The resulting $\rho_u$ and $\bar{A}_u$ are then used in the LoGo-GRPO objective.

\subsection{LoGo-GRPO for Multi-session Credit Assignment}
\label{sec:logo_grpo}

The formulation in Sec.~\ref{sec:problem_formulation} defines memory construction as a chunk-wise multi-step process, but learning still requires fair credit assignment across sessions. Trajectory-level GRPO is problematic in memory-augmented settings because memory turns an agent's past actions into part of its future environment. Once rollouts write, update, or delete different memories, they no longer share the same intermediate memory state, making group-relative comparisons unfair and credit signals noisy or biased. To address this, we propose \textbf{LoGo-GRPO}, which combines a global trajectory-level branch with a local rerollout branch. As shown in Figure~\ref{fig:teaser} (b), the global branch preserves end-to-end optimization over the full multi-session trajectory, while the local branch rerolls a stochastically sampled subset of sessions from shared memory states, yielding lower-bias session-level credit assignment at manageable cost.

\paragraph{Reward function.}
Let $\mathcal{Q}$ denote the full set of question-answer pairs $(q,a^*)$ associated with a training conversation, and let $\mathcal{Q}_t \subseteq \mathcal{Q}$ denote the subset whose required evidence is attributed to session $t$. Given a memory bank $\mathcal{M}$ and a question $q$, an answer module retrieves relevant entries from memory and generates an answer $\hat{a}$. We measure QA quality using token-level F1:
\begin{equation}
\mathrm{QA}(\mathcal{M}, \mathcal{Q}_t)
=
\frac{1}{|\mathcal{Q}_t|}
\sum_{(q,a^*) \in \mathcal{Q}_t}
\mathrm{F1}\!\left(\hat{a}, a^*\right).
\label{eq:qa_score}
\end{equation}
To discourage unbounded memory growth, we penalize memory tokens exceeding an $\alpha$ fraction of the cumulative session tokens up to session $t$, where $\mathrm{Tok}(\cdot)$ denotes token count and $\alpha$ is a fixed memory budget ratio:
\begin{equation}
\mathrm{Comp}(\mathcal{M}, t)
=
\begin{cases}
0,
& \mathrm{Tok}(\mathcal{M}) \le \alpha \sum_{s=1}^{t}\mathrm{Tok}(S_s), \\[6pt]
\dfrac{
\mathrm{Tok}(\mathcal{M}) - \alpha \sum_{s=1}^{t}\mathrm{Tok}(S_s)
}{
\sum_{s=1}^{t}\mathrm{Tok}(S_s)
},
& \mathrm{Tok}(\mathcal{M}) > \alpha \sum_{s=1}^{t}\mathrm{Tok}(S_s).
\end{cases}
\label{eq:compression_ratio}
\end{equation}
The session-level reward is
\begin{equation}
R(\mathcal{M}, \mathcal{Q}_t, t)
=
\mathrm{QA}(\mathcal{M}, \mathcal{Q}_t)
-
\lambda_{\mathrm{comp}}\,\mathrm{Comp}(\mathcal{M}, t),
\label{eq:reward}
\end{equation}
where $\lambda_{\mathrm{comp}}$ controls the compression penalty.

\paragraph{Global branch.}
For rollout $i$, let $\mathcal{M}_{t}^{(i)} \equiv \mathcal{M}_{t,K}^{(i)}$ denote the memory state after session $t$. The global branch evaluates the terminal memory $\mathcal{M}_{T}^{(i)}$ and attributes the reward to session $t$ according to the location of the required evidence:
\begin{equation}
r_{t,i}^{\mathrm{G}}
=
R\!\left(\mathcal{M}_{T}^{(i)},\, \mathcal{Q}_{t},\, T\right).
\label{eq:global_reward}
\end{equation}
Following GRPO, we compute group-relative advantages across the $n$ global rollouts:
\begin{equation}
\hat{A}_{t,i}^{\mathrm{G}}
=
\frac{r_{t,i}^{\mathrm{G}} - \mu_{t}^{\mathrm{G}}}
{\sigma_{t}^{\mathrm{G}} + \varepsilon},
\qquad
\mu_{t}^{\mathrm{G}}
=
\frac{1}{n}\sum_{j=1}^{n} r_{t,j}^{\mathrm{G}},
\qquad
\sigma_{t}^{\mathrm{G}}
=
\mathrm{std}_{j}\!\left(r_{t,j}^{\mathrm{G}}\right).
\label{eq:global_adv}
\end{equation}
While this branch provides full-horizon supervision, it still suffers from reward contamination: at session $t$, different rollouts induce different intermediate memory states as their effective environments, yet GRPO normalizes their rewards within the same comparison group.

\paragraph{Local branch with stochastic rerollout.}
To reduce this contamination, the local branch performs rerollouts from shared intermediate memory states. After the global rollout phase, each session is independently selected with probability $p_{\mathrm{local}}$:
\begin{equation}
b_t \sim \mathrm{Bernoulli}(p_{\mathrm{local}}),
\qquad
\mathcal{B} = \{ t \mid b_t = 1 \}.
\label{eq:local_buffer}
\end{equation}
For each selected session $t \in \mathcal{B}$, we choose an anchor rollout $i_0 \in \{1,\dots,n\}$, retrieve the cached memory state immediately before session $t$, and sample $m$ local rerollouts of session $t$ only. Since these rerollouts share the same starting memory state $\mathcal{M}_{t-1}^{(i_0)}$, their comparison is not confounded by divergence from earlier sessions. Let $\mathcal{M}_{t}^{(i_0,j)}$ denote the memory state after the $j$-th local rerollout from this anchor state. The local reward is
\begin{equation}
r_{t,j}^{\mathrm{L}}
=
R\!\left(\mathcal{M}_{t}^{(i_0,j)},\, \mathcal{Q}_{t},\, t\right),
\qquad j=1,\dots,m.
\label{eq:local_reward}
\end{equation}
The corresponding local advantages are computed within the rerollout group:
\begin{equation}
\hat{A}_{t,j}^{\mathrm{L}}
=
\frac{r_{t,j}^{\mathrm{L}} - \mu_{t}^{\mathrm{L}}}
{\sigma_{t}^{\mathrm{L}} + \varepsilon},
\qquad
\mu_{t}^{\mathrm{L}}
=
\frac{1}{m}\sum_{j=1}^{m} r_{t,j}^{\mathrm{L}},
\qquad
\sigma_{t}^{\mathrm{L}}
=
\mathrm{std}_{j}\!\left(r_{t,j}^{\mathrm{L}}\right).
\label{eq:local_adv}
\end{equation}
Because local advantages are computed among rerollouts from the same anchor memory state $\mathcal{M}_{t-1}^{(i_0)}$, the comparison is fairer than global normalization across already-diverged trajectories.

\paragraph{Unified training objective.}
We optimize the shared memory policy using both global rollouts and local rerollouts. For each generation step $u$, we assign the normalized advantage associated with its corresponding rollout: $\hat{A}_{t,i}^{\mathrm{G}}$ for a step from global rollout $i$ at session $t$, and $\hat{A}_{t,j}^{\mathrm{L}}$ for a step from local rerollout $j$ at session $t$. The same assigned advantage is used as the token-level advantage $A_{\ell}$ for all tokens in step $u$. Let $\mathcal{K}_{\mathrm{step}}$ denote the set of valid generation steps from both branches. Using the step-level ratio $\rho_u$ and advantage $\bar{A}_u$ from Eq.~\ref{eq:step_level}, we optimize the dual-clipped surrogate

\begin{equation}
\ell_u
=
\begin{cases}
\min\!\left(
-c\,\bar{A}_u,\;
\max\!\left(
-\rho_u \bar{A}_u,\;
-\mathrm{clip}(\rho_u,1-\epsilon,1+\epsilon)\bar{A}_u
\right)
\right),
& \bar{A}_u < 0, \\[2mm]
\max\!\left(
-\rho_u \bar{A}_u,\;
-\mathrm{clip}(\rho_u,1-\epsilon,1+\epsilon)\bar{A}_u
\right),
& \bar{A}_u \ge 0 ,
\end{cases}
\label{eq:step_dual_clip}
\end{equation}
where $c>1$ is the dual-clipping constant and $\epsilon$ is the clipping threshold. The final actor objective is
\begin{equation}
\mathcal{L}(\theta)
=
\frac{1}{|\mathcal{K}_{\mathrm{step}}|}
\sum_{u \in \mathcal{K}_{\mathrm{step}}} \ell_u
-
\beta_{\mathrm{ent}}\,\overline{H}_{\mathrm{token}}
+
\beta_{\mathrm{kl}}\,\overline{D}_{\mathrm{KL,token}},
\label{eq:unified_objective}
\end{equation}
where $\overline{H}_{\mathrm{token}}$ and $\overline{D}_{\mathrm{KL,token}}$ denote the mean token-level entropy and KL divergence, respectively. The proportion of local rerollouts controls the strength of local supervision, allowing LoGo-GRPO to balance end-to-end long-horizon learning with lower-bias session-level credit assignment.

\subsection{Curriculum Learning for Long-Horizon Credit Assignment}
Directly training on long multi-session trajectories is unstable before the model acquires reliable memory manipulation skills. Because memory operations shape the future environment, early insert, update, or delete errors can propagate across sessions and make long-horizon credit assignment increasingly noisy. We therefore adopt a curriculum over session horizon: training starts from shorter sessions, where memory effects are easier to observe and attribute, and gradually increases the horizon as the policy stabilizes. Concretely, we train in three stages with the maximum number of sessions increasing from 8 to 16 to 32. The 8-session stage learns basic memory operations under limited error propagation, the 16-session stage introduces stronger inter-session dependencies, and the 32-session stage enables full long-horizon optimization. For each stage, we select the best validation checkpoint as the initialization for the next stage, providing a stable starting point for longer-horizon training.


\section{Experiments}

\subsection{Experiment Setup}

\paragraph{Datasets and Evaluation Metrics.}
We train on LoCoMo~\cite{locomo-maharana2024evaluatinglongtermconversationalmemory}, a long-term persona-grounded conversation benchmark, using a 2:1:7 train/validation/test split. For out-of-distribution evaluation, we additionally test on LongMemEval~\cite{longmemeval-wu2025longmemevalbenchmarkingchatassistants}, MSC-Self-Instruct~\cite{mscselfinstruct-packer2024memgptllmsoperatingsystems,msc-xu-etal-2022-beyond}, and MemBench~\cite{membench-tan2025membenchcomprehensiveevaluationmemory}. We report token-level F1, BLEU-1 (B1), and LLM-as-a-Judge (J) as the primary metrics, and additionally use \textbf{M-Fail}, the percentage of required evidence location IDs that are missing from the memory bank, as a diagnostic measure of memory-construction quality. Further details on the M-Fail metric can be found in Appendix~\ref{app:evaluation_metrics}.

\paragraph{Baselines and Implementation Details.}
We compare against A-MEM~\cite{amem-xu2025amemagenticmemoryllm}, Mem0~\cite{mem0-chhikara2025mem0}, MemoryOS~\cite{memoryos-kang2025memoryosaiagent}, a RAG variant implemented within the Mem0 framework, MEM1~\cite{mem1-zhou2025mem1learningsynergizememory}, MemAgent~\cite{memagent-yu2025memagentreshapinglongcontextllm}, and Memory-R1~\cite{memoryr1-yan2025memory}. Our work primarily targets the memory construction stage: the memory extractor and memory manager share a Qwen2.5-7B-Instruct backbone and are jointly trained, while the answer agent is held fixed during training to provide stable reward signals. We use GPT-OSS-120B as this fixed answer agent, since a weaker answer model would yield noisy reward signals that conflate memory-construction quality with answer-generation errors. To remain consistent with this training pipeline, all reported results in our ablation and analysis experiments use the same GPT-OSS-120B answer agent. In Table~\ref{tab:main}, however, we additionally train a Qwen2.5-7B-Instruct answer agent and report a backbone-controlled variant of Memory-R2 in which all components share the same 7B backbone, enabling a fair comparison against the baselines. Unless otherwise noted, all baselines also use Qwen2.5-7B-Instruct as the backbone. Additional details are provided in Appendix~\ref{app:implementation_details}.

\begin{table}[t]
  \caption{Main results on LoCoMo. We report token-level F1 (F1), BLEU-1 (B1), and LLM-as-a-Judge (J), with the best per column in \textbf{bold}. For fair comparison, all baselines and Memory-R2 use Qwen2.5-7B-Instruct as the base model. We additionally report Memory-R2 (GPT-OSS), which swaps the answer agent for GPT-OSS-120B; notably, our RL-finetuned 7B Memory-R2 surpasses this 120B variant on F1 and B1, showing that targeted training outweighs raw model scale. Results are averaged over three runs; standard deviations are in Table~\ref{tab:main_result_variance}. $\dagger$: as reported in \cite{memoryr1-yan2025memory}.}
  \label{tab:main}
  \centering
  \resizebox{\textwidth}{!}{
  \begin{tabular}{l ccc ccc ccc ccc ccc}
    \toprule
    & \multicolumn{3}{c}{\textbf{Multi-hop}} & \multicolumn{3}{c}{\textbf{Open Domain}} & \multicolumn{3}{c}{\textbf{Single Hop}} & \multicolumn{3}{c}{\textbf{Temporal}} & \multicolumn{3}{c}{\textbf{Overall}} \\
    \cmidrule(r){2-4}\cmidrule(r){5-7}\cmidrule(r){8-10}\cmidrule(r){11-13}\cmidrule(r){14-16}
    \textbf{Model} & F1 & B1 & J & F1 & B1 & J & F1 & B1 & J & F1 & B1 & J & F1 & B1 & J \\
    \midrule
    \multicolumn{16}{l}{\emph{Training-free Methods}} \\
    RAG$^\dagger$                                              & 9.57 & 7.00 & 15.06 & 11.84 & 10.02 & 19.28 & 8.67 & 6.52 & 12.79 & 8.35 & 8.74 & 5.43 & 8.97 & 7.27 & 12.17 \\
    A-MEM~\cite{amem-xu2025amemagenticmemoryllm}$^\dagger$     & 18.92 & 12.86 & 40.78 & 14.73 & 12.66 & 31.32 & 30.58 & 26.14 & 46.90 & 23.67 & 20.67 & 28.68 & 26.08 & 21.78 & 40.78 \\
    Mem0~\cite{mem0-chhikara2025mem0}$^\dagger$                & 24.96 & 18.05 & 61.92 & 20.31 & 15.82 & 48.19 & 32.74 & 25.27 & 65.20 & 33.16 & 26.28 & 38.76 & 30.61 & 23.55 & 53.30 \\
    MemoryOS~\cite{memoryos-kang2025memoryosaiagent}$^\dagger$ & 29.55 & 22.59 & 48.12 & 21.03 & 18.41 & 38.55 & 40.85 & 36.26 & 63.14 & 26.26 & 19.70 & 24.81 & 34.64 & 29.36 & 51.26 \\
    \midrule
    \multicolumn{16}{l}{\emph{Trained Methods}} \\
    MEM1 \cite{mem1-zhou2025mem1learningsynergizememory} & 17.15 & 12.72 & 41.29 & 22.67 & 14.72 & 43.10 & 28.05 & 22.96 & 57.22 & 31.77 & 26.40 & 44.73 & 26.55 & 21.38 & 50.78 \\
    MemAgent \cite{memagent-yu2025memagentreshapinglongcontextllm}               & 35.95 & 24.56 & 70.65  & 29.20 & 25.48 & 62.07 & 50.55 & 43.89 & 78.61  & 23.15 & 16.55 & 56.96  & 40.72 & 33.36 & 71.52 \\
    Memory-R1~\cite{memoryr1-yan2025memory}$^\dagger$          & 33.64 & 26.06 & 62.34 & 23.55 & 20.71 & 40.96 & 46.86 & 40.92 & 67.81 & 47.75 & 38.49 & 49.61 & 43.14 & 36.44 & 61.51 \\
    \midrule
    Memory-R2(OSS)                                             & 36.37 & 29.79 & \textbf{85.07} & \textbf{30.22} & \textbf{24.50} & \textbf{74.14} & 50.98 & 45.79 & \textbf{90.58} & \textbf{62.48} & \textbf{55.33} & \textbf{83.33} & 49.67 & 43.77 & \textbf{87.10} \\
    Memory-R2                                                  & \textbf{38.41} & \textbf{30.90} & 80.93 & 20.76 & 16.78 & 67.53 & \textbf{54.06} & \textbf{48.73} & 86.80 & 59.65 & 50.05 & 69.90 & \textbf{50.60} & \textbf{44.01} & 80.99 \\
    \bottomrule
  \end{tabular}
  }
\end{table}

\begin{figure}[t]
  \centering
  \includegraphics[width=1\textwidth]{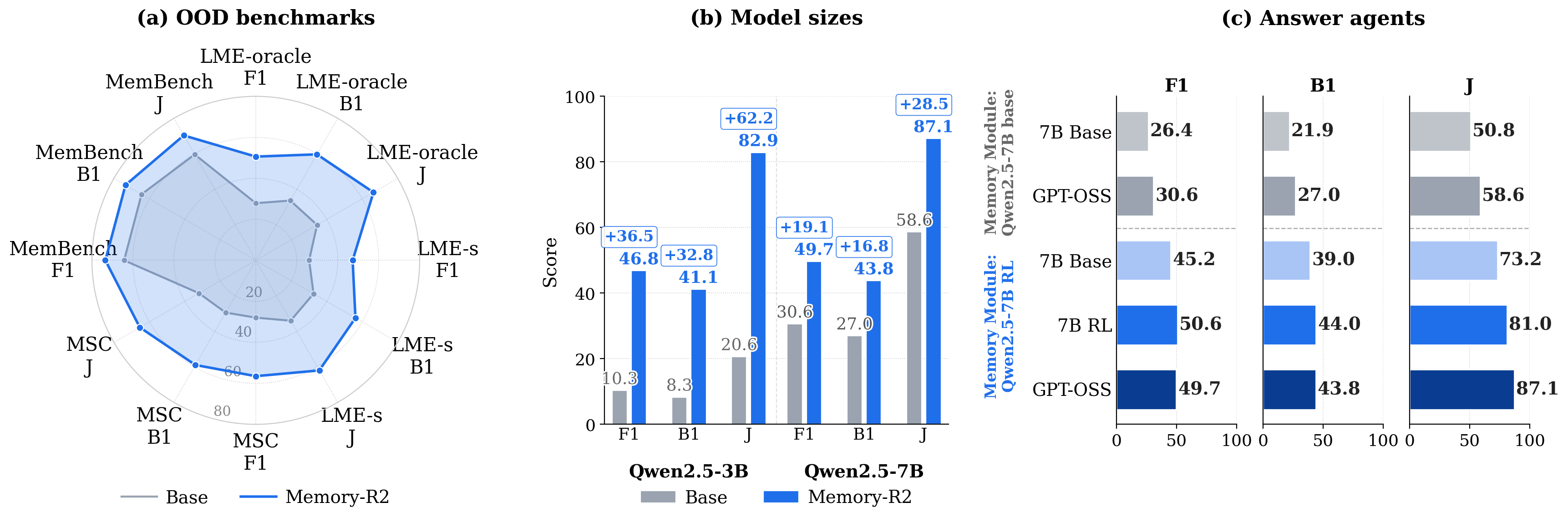}
  \caption{Generalization of Memory-R2 across (a) OOD benchmarks, (b) backbone sizes, and (c) answer agents.}
  \label{fig:generalization}
\end{figure}

\subsection{Main Results}
\label{sec:main_results}
\paragraph{Fair Comparison.}
Table~\ref{tab:main} reports the main results on LoCoMo. Under the backbone-controlled setting, Memory-R2 achieves the best overall F1 and BLEU-1 among all training-free and trained baselines, including MEM1, MemAgent, and Memory-R1. Compared with the closely related RL baseline Memory-R1, Memory-R2 improves overall F1 from 43.14 to 50.60 and B1 from 36.44 to 44.01, while also reaching a strong judge score of 80.99. These gains are obtained with a simple memory-agent pipeline, suggesting that the improvement mainly comes from the proposed training algorithm rather than additional system complexity. We additionally report Memory-R2 (GPT-OSS), which uses the same memory construction module but replaces the answer agent with GPT-OSS-120B. Memory-R2 with the 7B answer agent achieves higher F1 and BLEU-1 than the GPT-OSS-120B variant, demonstrating that a task-aligned small model can rival a much larger frozen one when paired with a well-trained memory module.

\paragraph{Strong Generalization.}
Figure~\ref{fig:generalization} further demonstrates the strong generalization ability of Memory-R2 from three complementary perspectives. Notably, these gains are achieved even though the model is trained on only two LoCoMo conversations, suggesting that the proposed training paradigm is highly data-efficient. First, Figure~\ref{fig:generalization}(a) shows strong transfer to out-of-distribution benchmarks. When evaluated zero-shot on LongMemEval-oracle, LongMemEval-s, MSC-Self-Instruct, and MemBench, Memory-R2 consistently improves over the base model across all reported metrics. For example, on LongMemEval-oracle, the F1 score improves from 27.88 to 50.60, and similar gains are observed on the other benchmarks, indicating that the learned memory-construction policy does not simply overfit to the training benchmark. Second, Figure~\ref{fig:generalization}(b) shows that the gains also transfer across model scales. The improvement is especially pronounced for Qwen2.5-3B, where F1 increases from 10.3 to 46.8, suggesting that our training paradigm is particularly beneficial for smaller-capacity models, for which effective long-horizon memory construction is otherwise difficult to learn. Third, Figure~\ref{fig:generalization}(c) decomposes the contribution of training the memory module versus the answer agent. The dominant gain comes from training the memory module (e.g., F1 from 26.4 to 45.2 with a 7B-Base answer agent; F1 from 30.6 to 49.7 with a GPT-OSS answer agent), while varying the answer agent at fixed RL-trained memory yields comparably high scores. This indicates that the benefits of Memory-R2 transfer across diverse downstream answer agents. Taken together, these results indicate that Memory-R2 learns a robust and transferable memory-construction policy rather than overfitting to a specific benchmark, model scale, or answer agent.

\begin{figure}[t]
  \centering
  \includegraphics[width=1\textwidth]{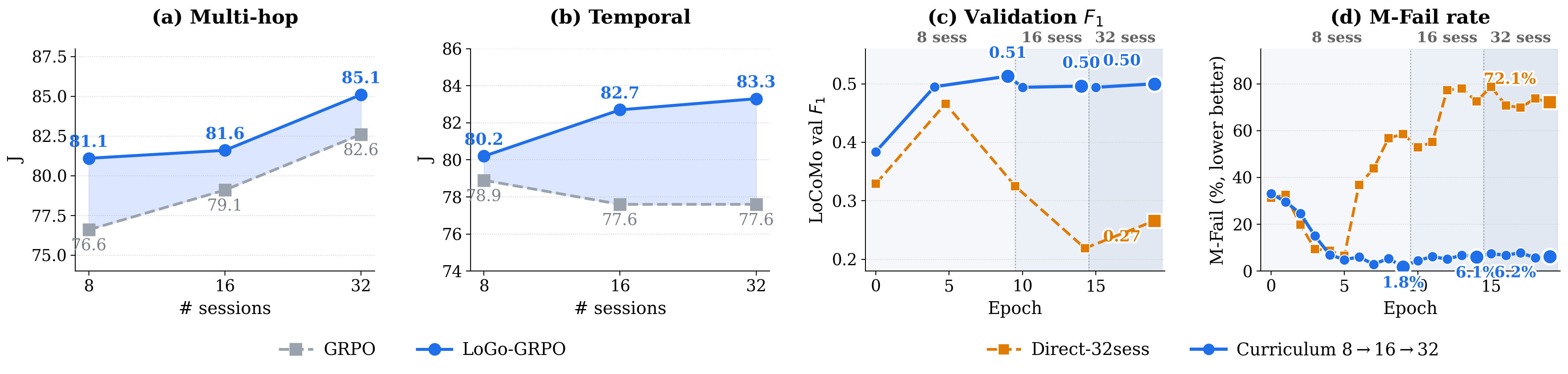}
   \caption{\textbf{LoGo-GRPO and curriculum learning are both essential.}
   \textbf{(a,b)} LoGo-GRPO consistently outperforms GRPO across curriculum stages.
   \textbf{(c,d)} Curriculum training remains stable under equal compute, whereas direct 32-session training collapses validation F1 from $0.47$ to $0.27$ and increases M-Fail to $72.1\%$.}
  \label{fig:logo_grpo_curriculum}
\end{figure}

\begin{wraptable}{r}{0.5\textwidth}
  \vspace{-1em}
  \caption{Ablation studies on components of LoGo-GRPO.}
  \label{tab:ablation}
  \centering
  \renewcommand{\arraystretch}{1.4}
  \setlength{\tabcolsep}{3pt}
  \footnotesize
  \begin{tabular}{lcccc}
    \toprule
    \textbf{Variants} & \textbf{F1} $\uparrow$ & \textbf{B1} $\uparrow$ & \textbf{J} $\uparrow$ & \textbf{M-Fail} $\downarrow$ \\
    \midrule
    \textbf{LoGo-GRPO (full)} & 49.67 & 43.77 & 87.10 & 6.72 \\
    \midrule
    \multicolumn{5}{l}{\textit{RL Algorithm Ablation}} \\
    \quad GRPO
      & \shortstack{46.62\\{\scriptsize\textcolor{red}{(-3.05)}}}
      & \shortstack{40.97\\{\scriptsize\textcolor{red}{(-2.80)}}}
      & \shortstack{82.76\\{\scriptsize\textcolor{red}{(-4.34)}}}
      & \shortstack{10.20\\{\scriptsize\textcolor{red}{(+3.48)}}} \\
    \quad -curriculum
      & \shortstack{24.12\\{\scriptsize\textcolor{red}{(-25.55)}}}
      & \shortstack{20.67\\{\scriptsize\textcolor{red}{(-23.10)}}}
      & \shortstack{45.99\\{\scriptsize\textcolor{red}{(-41.11)}}}
      & \shortstack{46.50\\{\scriptsize\textcolor{red}{(+39.78)}}} \\
    \quad -length\ norm.
      & \shortstack{43.53\\{\scriptsize\textcolor{red}{(-6.14)}}}
      & \shortstack{38.10\\{\scriptsize\textcolor{red}{(-5.67)}}}
      & \shortstack{77.20\\{\scriptsize\textcolor{red}{(-9.90)}}}
      & \shortstack{8.30\\{\scriptsize\textcolor{red}{(+1.58)}}} \\
    \midrule
    \multicolumn{5}{l}{\textit{Architecture Ablation}} \\
    \quad Single agent
      & \shortstack{39.14\\{\scriptsize\textcolor{red}{(-10.53)}}}
      & \shortstack{33.93\\{\scriptsize\textcolor{red}{(-9.84)}}}
      & \shortstack{72.63\\{\scriptsize\textcolor{red}{(-14.47)}}}
      & \shortstack{18.00\\{\scriptsize\textcolor{red}{(+11.28)}}} \\
    \quad Separate params
      & \shortstack{44.31\\{\scriptsize\textcolor{red}{(-5.36)}}}
      & \shortstack{38.59\\{\scriptsize\textcolor{red}{(-5.18)}}}
      & \shortstack{78.89\\{\scriptsize\textcolor{red}{(-8.21)}}}
      & \shortstack{33.00\\{\scriptsize\textcolor{red}{(+26.28)}}} \\
    \midrule
    \multicolumn{5}{l}{\textit{Chunks Ablation}} \\
    \quad N = 4
      & \shortstack{40.39\\{\scriptsize\textcolor{red}{(-9.28)}}}
      & \shortstack{35.16\\{\scriptsize\textcolor{red}{(-8.61)}}}
      & \shortstack{73.92\\{\scriptsize\textcolor{red}{(-13.18)}}}
      & \shortstack{14.50\\{\scriptsize\textcolor{red}{(+7.78)}}} \\
    \quad N = 8
      & \shortstack{41.37\\{\scriptsize\textcolor{red}{(-8.30)}}}
      & \shortstack{36.15\\{\scriptsize\textcolor{red}{(-7.62)}}}
      & \shortstack{75.58\\{\scriptsize\textcolor{red}{(-11.52)}}}
      & \shortstack{22.70\\{\scriptsize\textcolor{red}{(+15.98)}}} \\
    \quad N = 10
      & \shortstack{37.61\\{\scriptsize\textcolor{red}{(-12.06)}}}
      & \shortstack{32.61\\{\scriptsize\textcolor{red}{(-11.16)}}}
      & \shortstack{70.05\\{\scriptsize\textcolor{red}{(-17.05)}}}
      & \shortstack{37.20\\{\scriptsize\textcolor{red}{(+30.48)}}} \\
    \midrule
    \multicolumn{5}{l}{\textit{Training Target Ablation}} \\
    \quad memory manager
      & \shortstack{45.34\\{\scriptsize\textcolor{red}{(-4.33)}}}
      & \shortstack{39.99\\{\scriptsize\textcolor{red}{(-3.78)}}}
      & \shortstack{81.20\\{\scriptsize\textcolor{red}{(-5.90)}}}
      & \shortstack{16.10\\{\scriptsize\textcolor{red}{(+9.38)}}} \\
    \quad fact extractor
      & \shortstack{28.30\\{\scriptsize\textcolor{red}{(-21.37)}}}
      & \shortstack{24.41\\{\scriptsize\textcolor{red}{(-19.36)}}}
      & \shortstack{52.90\\{\scriptsize\textcolor{red}{(-34.20)}}}
      & \shortstack{56.50\\{\scriptsize\textcolor{red}{(+49.78)}}} \\
    \bottomrule
  \end{tabular}
  \renewcommand{\arraystretch}{1}
  \vspace{-0.5em}
\end{wraptable}

\subsection{Ablation Studies}
\label{sec:ablation}

Table~\ref{tab:ablation} summarizes ablations on the major components of our method, with \textbf{M-Fail} reported as a diagnostic measure of memory quality.

Replacing LoGo-GRPO with standard GRPO degrades F1 from 49.67 to 46.62 and B1 from 43.77 to 40.97, confirming the benefit of global-local credit assignment. Figure~\ref{fig:logo_grpo_curriculum}(a, b) shows this gap holds at every stage across question types, indicating that local rerollouts consistently mitigate credit-assignment bias. Removing curriculum learning ($-$curriculum) causes a much larger drop—F1 falls to 24.12 and M-Fail rises to 46.5\%. Figure~\ref{fig:logo_grpo_curriculum}(c, d) traces this collapse: direct 32-session training peaks at $F_1{=}0.47$ before falling to $0.27$, while M-Fail explodes from below 10\% to over 70\%; the curriculum instead stabilizes around $F_1{=}0.50$ with M-Fail held under 7\%. This confirms that early errors propagate across sessions and corrupt memory, and that progressive horizon expansion is essential for stable long-horizon training. We further ablate the length normalization in our step-level objective: switching to a token-level loss ($-$length norm.) drops F1 to 43.53 and B1 to 38.10, confirming that length-normalized step weighting is necessary to prevent output length bias under the shared extractor--manager policy.

For the memory-construction architecture, a single-agent variant merging extraction and editing into one role drops to 39.14 F1, and a separate-params variant where the extractor and manager use disjoint parameters also underperforms (44.31 F1), supporting both explicit role decomposition and parameter sharing. Alternative interaction depths likewise underperform the full multi-step design (40.39 / 41.37 / 37.61 F1 for $N{=}4/8/10$), showing that moderate iterative refinement is optimal. Too few chunks limit refinement, while overly long interaction chains hurt optimization. Finally, training only the memory manager (45.34 F1) or only the fact extractor (28.30 F1) also degrades performance, especially the latter, confirming that both components benefit from joint RL training and that fact extraction is the more brittle of the two roles when left untrained.

Overall, the gains of our method arise from the combination of fair credit assignment, curriculum learning, multi-step memory construction, and shared extractor--manager co-learning. Additional ablations are reported in Appendix~\ref{app:additonal_ablations}.

\subsection{More Analysis: Latency and Compression}
\label{sec:more_analysis}

\begin{figure}[t]
  \centering
  \includegraphics[width=1\textwidth]{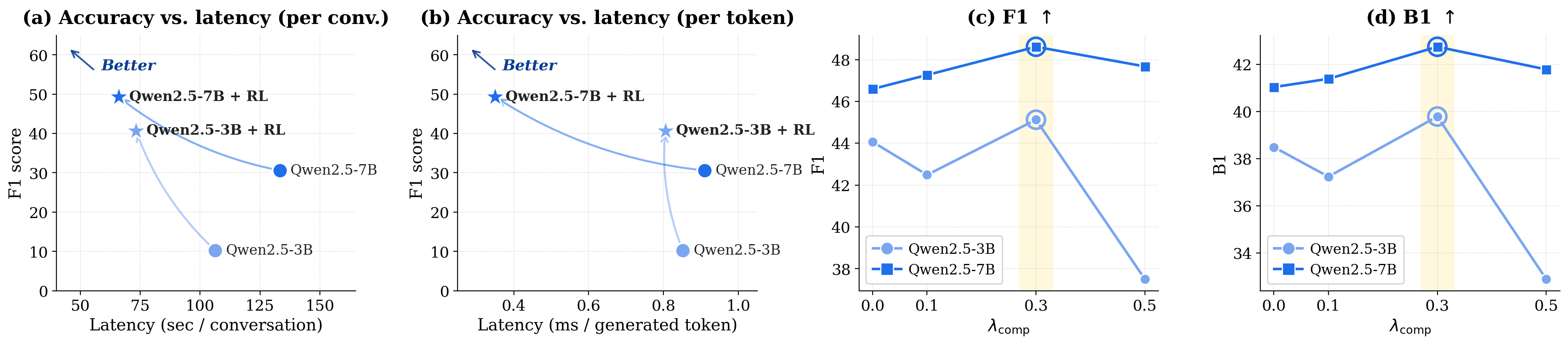}
  \caption{\textbf{Inference efficiency and compression penalty analysis.}
\textbf{(a,b)} Accuracy--latency trade-off measured by F1 vs.\ time per conversation and per generated token.
\textbf{(c,d)} Effect of $\lambda_{\mathrm{comp}} \in \{0,0.1,0.3,0.5\}$ on F1 and BLEU-1; the yellow band marks $\lambda_{\mathrm{comp}}=0.3$, and rings mark the best value.}
  \label{fig:latency_compression}
\end{figure}

\paragraph{Latency.}
Figure~\ref{fig:latency_compression}(a,b) compares F1 and inference latency before and after Memory-R2 training. Memory-R2 improves F1 while reducing latency for both Qwen2.5-3B and Qwen2.5-7B under the per-conversation measurement, moving both models toward a better quality--efficiency regime. The source of this latency reduction differs across scales. For Qwen2.5-3B, the gain is mainly driven by more concise generations: the trained policy emits fewer tokens per memory-construction turn. For Qwen2.5-7B, the gain comes not only from shorter generations, but also from a more stable generation-length distribution: the untrained policy occasionally produces overly long memory-management outputs, whereas Memory-R2 suppresses these unnecessary generations, reducing decoding work and making memory construction more stable. We provide a diagnostic breakdown of these scale-dependent mechanisms in Figure~\ref{fig:latency_mechanism}. These results suggest that better-trained memory policies can improve answer quality without incurring additional inference overhead, and can even reduce latency by making memory construction more concise and stable.

\paragraph{Compression.}
Figure~\ref{fig:latency_compression}(c,d) studies the effect of the compression penalty $\lambda_{\mathrm{comp}}$. Across both Qwen2.5-3B and Qwen2.5-7B, $\lambda_{\mathrm{comp}}=0.3$ achieves the best F1 and BLEU-1, as highlighted by the yellow band. Smaller penalties may retain redundant or noisy memories, while overly strong compression can remove useful evidence. We therefore use $\lambda_{\mathrm{comp}}=0.3$ as the default setting.

\section{Conclusion}

In this paper, we present Memory-R2, a training framework for long-horizon memory-augmented LLM agents that addresses a fundamental challenge in multi-session reinforcement learning: fair credit assignment under diverging memory states. Our method, LoGo-GRPO, combines global trajectory-level optimization with local rerollouts from shared intermediate memory states, enabling fairer session-level comparisons while preserving end-to-end long-horizon learning. Beyond credit assignment, Memory-R2 jointly optimizes memory formation and memory evolution through a shared extractor--manager policy, formulates memory construction as a multi-step decision process over chunked sessions, and stabilizes training with a curriculum over session horizon. Experiments show that Memory-R2 consistently outperforms prior memory-agent baselines on LoCoMo and generalizes well across out-of-distribution benchmarks, model scales, and answer agents. These results suggest that improving credit assignment is a key ingredient for training robust long-horizon memory agents, and we hope this work provides a useful foundation for future research on memory-centric RL for LLM agents.

\bibliographystyle{plainnat}
\bibliography{references}

\newpage
\appendix

\section{Additional Implementation Details}
\label{app:implementation_details}

\begin{figure}[!h]
  \centering
  \includegraphics[width=1\textwidth]{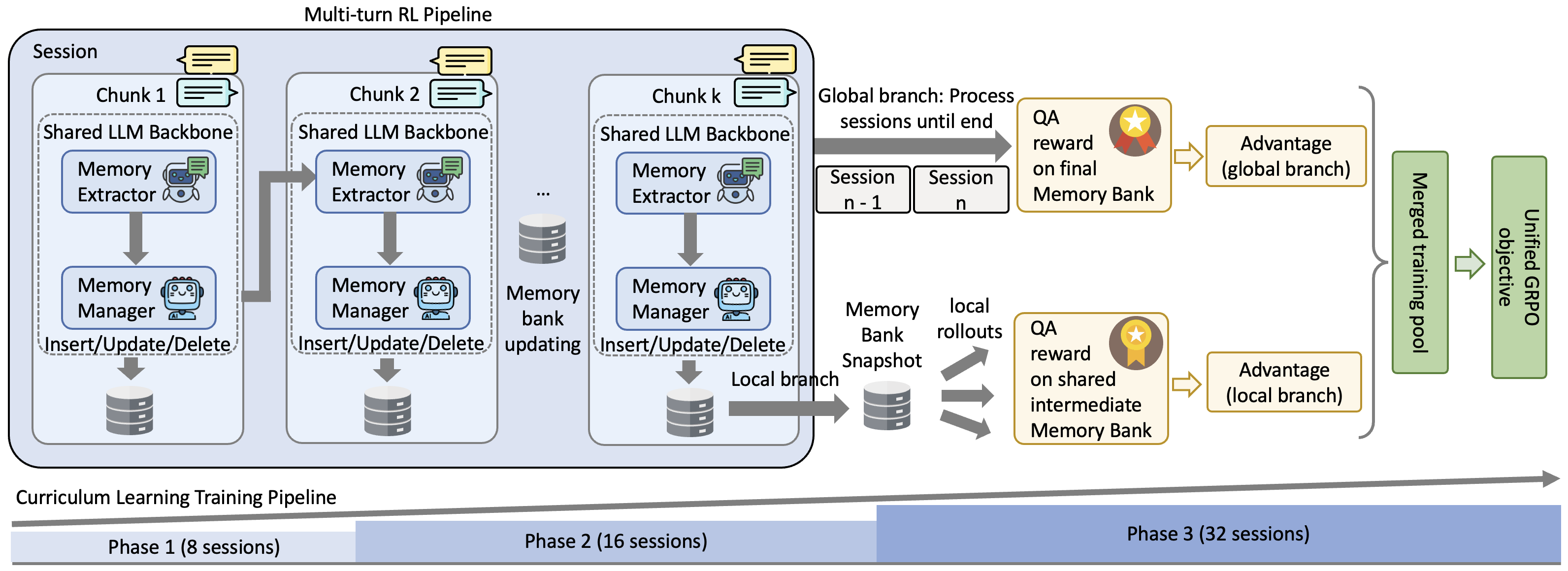}
  \caption{LoGo-GRPO training pipeline for memory manager. Memory bank construction via alternating extraction and management steps over chunked sessions. Global rollouts optimize end-to-end performance using rewards from the final memory, while local rerollouts from shared memory states provide low-bias credit assignment. Both signals are unified in a single GRPO-style objective, with curriculum learning enabling stable long-horizon training.}
  \label{fig:pipeline}
\end{figure}

\subsection{RL Training for Answer Agent}
\label{app:rl_training_for_answer_agent}
\textit{Training pipeline.}
We fine-tune Qwen2.5-7B-Instruct as the answer agent using GRPO, implemented with the VERL framework~\cite{verl-sheng2024hybridflow}. Each training example is a single-turn QA prompt. Given a natural-language question $q$ and a constructed memory bank $\mathcal{M}$, we retrieve, for each speaker, the top-$30$ memory entries by text-embedding similarity to $q$ using a similarity threshold of $0.3$, and insert them into the QA template (Appendix~\ref{app:answer-agent-prompt-template}). The template instructs the model to reason step by step over the timestamped memories and output the final answer inside an \texttt{<answer>...</answer>} tag.

\textit{Training data.}
We construct the training set by running the full memory-construction pipeline (fact extraction, memory operations, and QA answering) on only two LoCoMo training conversations with GPT-4o, $T{=}0$. For each QA pair $(q, a^\star)$, we store (i) the rendered QA prompt and (ii) the answer extracted from the generated \texttt{<answer>} tag. Because some QA instances in the original data are noisy or weakly aligned with the available memory evidence, we retain only samples whose generated answer achieves token-level F1 $\geq 0.25$ against the gold answer $a^\star$. This filtering removes clearly problematic QA instances while preserving a sufficiently diverse training distribution. The remaining samples are randomly split into $90\%$ training and $10\%$ validation.

\textit{Reward.}
We use a judge-free rule-based reward
\[
R(\hat{y}, a^\star) = \mathrm{F1}\bigl(\mathrm{extract}(\hat{y}),\, a^\star\bigr) \in [0,1],
\]
where $\mathrm{extract}(\cdot)$ returns the substring inside the first \texttt{<answer>...</answer>} tag. The F1 score is computed using the standard SQuAD-style token F1 after lowercasing, removing punctuation and articles (\texttt{a}, \texttt{an}, \texttt{the}), and whitespace tokenization. This keeps the training reward aligned with the answer-level metric used in evaluation.

\textit{Optimization.}
We train with GRPO using $n{=}8$ rollouts per prompt, sampling temperature $1.0$, the vLLM backend, GPU memory utilization $0.8$, and tensor parallelism $\mathrm{TP}{=}1$. We set the maximum prompt length to $12{,}288$ tokens and the maximum response length to $1{,}024$ tokens, with left truncation applied on the prompt side. The train batch size is $64$, the PPO mini-batch size is $16$, and the micro-batch size is $1$ per GPU. We use the standard GRPO advantage estimator and apply a token-level KL penalty in the actor loss with coefficient $0.001$, without adding KL to the reward. For runs with at least two GPUs, both parameters and optimizer states are kept on device; optimizer offloading is enabled only in the single-GPU setting. We train for $5$ epochs, and perform evaluation and checkpointing every $5$ optimizer steps.

\subsection{RL Training for Fact Extraction and Memory Management}
\label{app:rl_training_for_fact_extraction_and_memory_manager}

We train the joint fact-extractor and memory-manager agent using a curriculum RL recipe with VERL framework~\cite{verl-sheng2024hybridflow}. A single \texttt{Qwen2.5-7B-Instruct} backbone is shared across both roles: the fact extractor produces atomic facts, while the memory manager predicts \texttt{INSERT}/\texttt{UPDATE}/\texttt{DELETE} operations. Parameter sharing is realized through alternating role-conditioned rollouts within each session chunk.

\textit{Training Data.}
We use LoCoMo with a conversation-level 2:1:7 train/validation/test split. The memory-construction policy is trained using only the two conversations in the training split, which contain 328 associated QA pairs in total. These QA pairs are used to compute downstream rewards for memory construction, while the held-out validation conversation is used for checkpoint selection and the remaining seven conversations are reserved for test evaluation.

\textit{Optimization.}
We optimize the policy with GRPO~\citep{deepseekr1-Guo_2025}, using $N_{\text{rollout}}{=}16$ global trajectories per prompt and a local GRPO sampling fraction of $0.5$ with $N_{\text{local}}{=}4$ resampled turns to reduce gradient variance for memory operations. The actor is updated with $E_{\text{ppo}}{=}2$ epochs per batch, PPO mini-batch size $16$, micro-batch size $1$ per GPU, learning rate $\eta{=}2\times10^{-6}$, clipping ratio $\epsilon{=}0.2$, and entropy coefficient $0.001$. We use a KL penalty with coefficient $\beta_{\text{KL}}{=}10^{-3}$, without adding KL into the reward. We apply \emph{turn-level} importance-ratio clipping (\texttt{clip\_mode=turn}) and \emph{turn-level} loss aggregation.

\textit{Sequence and Rollout Budgets.}
Each turn uses a prompt budget of $L_{\text{prompt}}{=}28{,}672$ tokens and a response budget of $L_{\text{resp}}{=}4{,}096$ tokens. We cap the number of memory turns per session at $T_{\max}{=}4$ and stop early when generation is truncated. The rollout vLLM engine uses tensor parallelism $1$, GPU memory utilization $0.5$, and a maximum of $2(L_{\text{prompt}}{+}L_{\text{resp}})$ batched tokens per step.

\textit{Curriculum Learning.}
We adopt a session-length curriculum on LoCoMo. Stage 1 trains on trajectories truncated to 8 sessions for 10 epochs, followed by Stage 2 and Stage 3, which expand the horizon to 16 and 32 sessions and are trained for 5 epochs each.


\section{Prompt Templates}
\label{app:prompt-template}

\subsection{Prompt Template for Fact Extraction (Memory Formation)}
\label{app:fact-extraction-prompt}

The memory formation pipeline first extracts atomic, self-contained facts from raw dialogue turns before the memory manager integrates them
into the persistent store. Figure~\ref{fig:fact-extraction-prompt} shows
the prompt used to drive this fact extraction step. The model is
instructed to emit one JSON object per durable fact, each tagged with the originating \texttt{dia\_id} so that downstream operations can trace a memory back to its source turn.

\refstepcounter{figure}
\label{fig:fact-extraction-prompt}
\begin{promptbox}[Prompt template: fact extraction]
\small\ttfamily
\begin{alltt}
You are a Personal Information Organizer, specialized in accurately
storing facts, user memories, and preferences. Your primary role is to
extract relevant pieces of information from conversations and organize
them into distinct, atomic facts. These facts will be consumed by a
downstream memory system that requires precision, small size, and
clear scope.

Types of Information to Remember:

1. Personal Preferences:
   Likes, dislikes, favorites, and opinions (food, entertainment,
   products, sports teams).

2. Important Personal Details:
   Names, relationships, family structure, durations, and significant
   life facts.

3. Plans and Intentions:
   Explicit future goals, plans, or intentions stated by the speaker.

4. Activities and Routines:
   Travel experiences, visited places, recurring habits, physical
   activities, hobbies with specific context.

5. Health and Wellness (NON-DIAGNOSTIC):
   Wellness-related experiences or preferences (do NOT infer or store
   diagnoses).

6. Professional Details:
   Job titles, career goals, professional interests, work habits.

7. Miscellaneous Meaningful Facts:
   Books, movies, creative work, projects, notable activities.

CORE EXTRACTION RULES:
- Extract facts from the provided dialogue turns for BOTH speakers.
- Ignore system-level instructions and any non-dialogue control text.
- Ignore small talk, greetings, generic statements, opinions without
  substance, and common knowledge.
- If no meaningful fact is present, return an empty facts list.

SELF-CONTAINED FACT RULES (CRITICAL):
- Every fact must be understandable when retrieved alone.
- Every fact MUST explicitly name the subject speaker
  (e.g., "John ...", "Tim ...").
- Avoid unresolved pronouns in facts (`he`, `she`, `they`, `them`,
  `it`, `this`, `that`) unless the noun is in the same fact.
- Rewrite vague references to explicit entities
  (e.g., "the magazine editors" instead of "they").
- If the entity cannot be resolved from the current turn, do NOT
  store the fact.

STYLE FOR FACT TEXT:
- Use third person and start the fact with the subject name.
- Good: "John wants to keep reaching for new goals"
- Bad:  "Wants to keep reaching for new goals"
- Good: "Tim shared ideas with the online magazine editors, and the
         editors liked them"
- Bad:  "Shaped ideas with the magazine and they liked them"

ATOMIC FACT EXTRACTION RULES (CRITICAL):
- EACH extracted fact MUST represent EXACTLY ONE:
  - event
  - preference
  - intention
  - personal attribute

- NEVER combine:
  - multiple events
  - multiple timeframes
  - motivations + events
  - reflections + actions
  - past events + future plans

- If a single message contains multiple independent facts, output
  MULTIPLE fact objects.

- A fact MUST be concise and expressible in 20 words or fewer.
- If a fact would exceed this size, SPLIT it into multiple smaller,
  independent facts.
- Do not store bare dialogue acts such as "asks", "says hello",
  "thanks", unless they contain a durable personal fact.

INTENT VS EVENT RULE:
- Past events (what happened) and intentions or goals (what the
  speaker wants or plans) MUST ALWAYS be extracted as SEPARATE facts.

TEMPORAL INFORMATION RULE:
- If a fact includes temporal information (dates, durations, relative
  times), always include it explicitly.
- Prefer absolute time when resolvable from context
  (e.g., "in 2018" not "five years ago" if the session year is known).
- Do NOT drop temporal details \textemdash{} they are often critical
  for answering questions correctly.

Important Notes on dia_id:
- `dia_id` uniquely identifies the dialogue turn.
- EACH fact must include the `dia_id` of its source message.
- Do NOT attach multiple dia_ids to a single fact.

Here are some few-shot examples:

Input: [\{"speaker": "John", "text": "Hi, how are you?",
         "dia_id": "D1:1"\}]
Output: \{"facts": []\}

Input: [\{"speaker": "John", "text": "There are branches in trees.",
         "dia_id": "D2:3"\}]
Output: \{"facts": []\}

Input:
[
  \{"speaker": "Maria", "text": "What's your favorite sport?",
   "dia_id": "D3:1"\},
  \{"speaker": "John",  "text": "I love playing basketball with
                                  friends.", "dia_id": "D3:2"\}
]
Output:
\{"facts": [\{"speaker": "John", "dia_id": "D3:2",
             "fact": "John loves playing basketball with friends"\}]\}

Input:
[
  \{"speaker": "Maria", "text": "What did you do yesterday?",
   "dia_id": "D3:5"\},
  \{"speaker": "John",  "text": "Yesterday, I had a meeting at 3pm.
                                  We discussed a new project.",
   "dia_id": "D3:6"\}
]
Output:
\{"facts": [\{"speaker": "John", "dia_id": "D3:6",
             "fact": "John had a meeting at 3pm yesterday about a
                      new project"\}]\}

Input: [\{"speaker": "John", "text": "I am a software engineer.",
         "dia_id": "D4:2"\}]
Output:
\{"facts": [\{"speaker": "John", "dia_id": "D4:2",
             "fact": "John is a software engineer"\}]\}

Input: [\{"speaker": "John", "text": "My favorite movies are
         Inception and Interstellar.", "dia_id": "D4:3"\}]
Output:
\{"facts": [\{"speaker": "John", "dia_id": "D4:3",
             "fact": "John's favorite movies are Inception and
                      Interstellar"\}]\}

Input: [\{"speaker": "John", "text": "I attended an LGBTQ workshop
         last Friday and it inspired me to pursue counseling.",
         "dia_id": "D5:1"\}]
Output:
\{"facts": [
  \{"speaker": "John", "dia_id": "D5:1",
   "fact": "John attended an LGBTQ workshop last Friday"\},
  \{"speaker": "John", "dia_id": "D5:1",
   "fact": "John feels inspired to pursue a counseling career"\}
]\}

Input:
[
  \{"speaker": "Maria", "text": "That sounds wonderful!",
   "dia_id": "D6:3"\},
  \{"speaker": "John",  "text": "Thanks! I really appreciate your
                                support.", "dia_id": "D6:4"\}
]
Output: \{"facts": []\}

Input: [\{"speaker": "John", "text": "So what do you think about
         that?", "dia_id": "D7:2"\}]
Output: \{"facts": []\}

Return the facts in JSON format exactly as shown above.

Remember the following:
- If multiple statements describe the SAME EVENT at the SAME TIME
  and PLACE within the SAME dialogue turn, they MAY be merged into
  a SINGLE fact.
- If they differ by time, place, motivation, outcome, or reflection
  -> extract SEPARATE facts.
- Do NOT output standalone facts that depend on another fact for
  context unless merged into a complete event.
- Do not return anything from the custom few-shot example prompts
  provided above.
- If no relevant facts are found, return \{"facts": []\}.
- The response MUST be valid JSON with a single top-level key:
  "facts".
\end{alltt}
\end{promptbox}

\vspace{2pt}
\begin{flushleft}
{\small\textbf{Figure~\thefigure:} Prompt template for atomic fact
extraction. Each extracted fact is a self-contained, third-person
statement tagged with the originating \texttt{dia\_id}, and is then
passed to the memory manager (Figure~\ref{fig:memory-manager-prompt})
for integration into the persistent memory store.\par}
\end{flushleft}

\subsection{Prompt Template for the Memory Manager (Memory Evolution)}
\label{app:memory-manager-prompt}

The memory manager is the second stage of our pipeline. It takes the
atomic facts produced by the fact-retrieval prompt
(Section~\ref{app:fact-extraction-prompt}) together with the current
memory store, and decides for each new fact whether to insert,
update, delete, or take no operation on the
store. Compared with a naive memory writer, our prompt enforces three properties that proved important in practice: (i) \emph{atomicity}, so
that each memory entry encodes exactly one fact; (ii) \emph{monotonicity},
so that prior factual claims are never silently dropped during an
\textsc{Update}; and (iii) \emph{noise tolerance} over the embedding-based candidates, which can be topically
unrelated. Figure~\ref{fig:memory-manager-prompt} shows the full prompt.

\refstepcounter{figure}\label{fig:memory-manager-prompt}
\begin{promptbox}[Prompt template: memory bank management]
\small\ttfamily
\begin{alltt}
You are a smart memory manager which controls the memory of a system.
You can perform four operations: (1) insert into the memory, (2) update
the memory, (3) delete from the memory, and (4) no change.

Your primary goal is to preserve accurate factual evidence over time.
Memory updates must be SAFE, NON-DESTRUCTIVE, and FACT-PRESERVING.

CONTENT LENGTH RULE (enforced before all other rules):
- Every `content` field you write (INSERT or UPDATE) MUST be at most
  20 words.
- ONE fact per memory item -- never combine multiple independent
  facts into one entry.
- If you cannot express a fact in 20 words, write the most essential
  part only.

INPUT FORMAT: The input contains two sections:
- "memories": a flat list of existing memory entries (each appears
  exactly once, identified by memory_id).
- "facts": new facts to process, each with a "related_memory_ids"
  list -- IDs pointing into "memories" as candidates for UPDATE or
  DELETE for that specific fact.

To find candidates for a fact: look up its "related_memory_ids" in
the "memories" list by memory_id.
WARNING: "related_memory_ids" are retrieved by embedding similarity
and MAY CONTAIN NOISE -- some IDs may point to memories that are
topically unrelated to the fact. Do NOT blindly UPDATE or DELETE
just because an ID appears in "related_memory_ids". Always verify
the memory content actually refers to the same entity and topic
before acting on it. If no entry is genuinely relevant, treat
"related_memory_ids" as empty.

For each new fact, decide whether to:
- INSERT: The fact is new and not captured by any entry in its
  "related_memory_ids" (or list is empty or has no genuinely relevant
  entry).
- UPDATE: The fact refers to the SAME entity or event as a
  "related_memory_ids" entry and enriches, refines, or corrects it
  WITHOUT removing prior factual information.
- DELETE: The fact explicitly proves a "related_memory_ids" entry is
  false or invalid (not merely outdated).
- NO OPERATION: The fact is already captured by a "related_memory_ids"
  entry, redundant, irrelevant, or insignificant.

There are specific guidelines to select which operation to perform:

1. INSERT: If the fact contains new information not captured in its
   `related_memory_ids`, then you have to add it.
   - Assign `speaker` as who the fact is ABOUT.
   - Assign `content` as a concise summary in third person.
   - Keep tense faithful to the source fact (past events may stay
     past tense).
   - Do NOT assign `memory_id` for INSERT operations; the system
     will auto-generate it.
   - Always include the `dia_id` with each inserted fact to ensure
     the memory is accurately linked to the correct dialogue.
   - `content` must be SELF-CONTAINED and include the subject name
     explicitly (e.g., "John ...", "Tim ...").
   - `content` must avoid vague pronouns unless the referenced noun
     appears in the same sentence.
   - If a fact cannot be made self-contained without guessing, skip
     it (NO OPERATION).

   Example:
   - Input: \{
       "memories": [\{"memory_id": "a32b32c1", "speaker": "John",
         "content": "John works as a software engineer",
         "session_time": "6:59 pm on 26 August, 2023",
         "dia_ids": ["D1:4"]\}],
       "facts": [
         \{"speaker": "John", "dia_id": "D3:6",
          "fact": "John had a meeting at 3pm",
          "related_memory_ids": ["a32b32c1"]\},
         \{"speaker": "John", "dia_id": "D3:6",
          "fact": "John discussed a new project",
          "related_memory_ids": ["a32b32c1"]\}
       ]
     \}
   - Operations:
     \{"operations": [
        \{"operation": "INSERT", "speaker": "John",
         "content": "John had a meeting at 3pm about a new project",
         "dia_id": "D3:6"\}
     ]\}

ATOMICITY RULE:
- Each memory item MUST represent a single fact or event.
- Do NOT merge multiple independent facts into one memory item.
- If a new fact represents a genuinely different event, topic, or
  attribute -> INSERT instead of UPDATE.
- Exception: if the new fact is a progression or status change of
  the SAME entity's story (e.g., "exploring a job" -> "accepted the
  job"), UPDATE the existing entry rather than inserting a duplicate.

2. UPDATE: Use UPDATE only when the new fact clearly refers to the
   SAME entity or event as an entry in its `related_memory_ids` and
   ADDS detail, refinement, or correction WITHOUT removing prior
   facts.
   - NEVER remove existing factual information during an UPDATE.
   - If the new fact is more specific, merge it with the existing
     content.
   - If both convey the same meaning, keep the more informative
     version.
   - If the new fact introduces a completely unrelated event, goal,
     or topic -> INSERT instead.
   - If the new fact is a later development or confirmation of the
     SAME entity's ongoing story (e.g., plan -> outcome, exploring
     -> confirmed) -> UPDATE even if the time is different.
   - Please keep in mind while updating you have to use the same ID.
   - Always include the `dia_id` with each updated fact to ensure
     the memory is accurately linked to the correct dialogue.
   - Please note to return the IDs in the output from the input IDs
     only and do not generate any new ID.

   Example (refinement -- same entity, added detail):
   - Input: \{
       "memories": [\{"memory_id": "a0299e69", "speaker": "Emily",
         "content": "Likes to play cricket",
         "session_time": "2:04 pm on 3 September, 2021",
         "dia_ids": ["D5:2"]\}],
       "facts": [\{"speaker": "Emily",
         "fact": "Emily loves to play cricket with friends",
         "dia_id": "D5:4", "related_memory_ids": ["a0299e69"]\}]
     \}
   - Operations:
     \{"operations": [\{"operation": "UPDATE",
       "memory_id": "a0299e69",
       "content": "Emily loves to play cricket with friends",
       "dia_id": "D5:4"\}]\}

   Example (cross-session fact evolution -- status changed from
   exploring to confirmed):
   - Input: \{
       "memories": [\{"memory_id": "f3a91b44", "speaker": "Sarah",
         "content": "Sarah is exploring a job opportunity at a tech
                     company in Seattle",
         "session_time": "3:00 pm on 10 March, 2022",
         "dia_ids": ["D2:5"]\}],
       "facts": [\{"speaker": "Sarah", "dia_id": "D3:8",
         "fact": "Sarah accepted a senior software engineer role at
                  TechCorp in Seattle",
         "related_memory_ids": ["f3a91b44"]\}]
     \}
   - Operations:
     \{"operations": [\{"operation": "UPDATE",
       "memory_id": "f3a91b44",
       "content": "Sarah accepted a senior software engineer role at
                   TechCorp in Seattle",
       "dia_id": "D3:8"\}]\}
   Explanation: Same entity (Sarah's job in Seattle), status evolved
   from "exploring" to "accepted" -- UPDATE the same memory entry.
   Do NOT insert a duplicate about the Seattle job.

INVALID UPDATE EXAMPLE (DO NOT DO THIS):
- Memory content: "Sarah traveled to Paris and Rome on her European
  trip"
- Wrong UPDATE content: "Sarah traveled to Paris" -- removes Rome,
  destroys stored fact
- Correct action: NO OPERATION (no new information) or INSERT a
  separate fact about Rome if it was new.

3. DELETE: Use DELETE only when a new fact explicitly contradicts
   and invalidates an entry in its `related_memory_ids`.
   - Do NOT delete memories just because they are old or less
     relevant.
   - Please note to return the IDs in the output from the input IDs
     only and do not generate any new ID.

   Example:
   - Input: \{
       "memories": [\{"memory_id": "6v0k193d", "speaker": "Samy",
         "content": "I went to Paris last summer",
         "session_time": "8:04 am on 3 February, 2009",
         "dia_ids": ["D6:5"]\}],
       "facts": [\{"speaker": "Samy",
         "fact": "Samy never went to Paris", "dia_id": "D7:1",
         "related_memory_ids": ["6v0k193d"]\}]
     \}
   - Operations:
     \{"operations": [\{"operation": "DELETE",
       "memory_id": "6v0k193d"\}]\}

4. NO OPERATION: If the new fact is already captured by an entry in
   its `related_memory_ids` -- even if worded differently -- do NOT
   insert a new entry.
   Before deciding INSERT, look up the fact's `related_memory_ids`
   in "memories" and check for semantic overlap: same person, same
   topic, same meaning.
   If a semantically equivalent memory already exists -> NO
   OPERATION (not INSERT).
   If `related_memory_ids` is empty -> INSERT is safe.

   Example (exact match):
   - Input: \{
       "memories": [\{"memory_id": "9b3c82e0", "speaker": "Sofia",
         "content": "Sofia loves cheese pizza",
         "session_time": "11:10 am on 18 March, 2020",
         "dia_ids": ["D8:3"]\}],
       "facts": [\{"speaker": "Sofia",
         "fact": "Sofia loves cheese pizza", "dia_id": "D8:10",
         "related_memory_ids": ["9b3c82e0"]\}]
     \}
   - Operations: \{"operations": []\}

   Example (semantic match -- paraphrase is NOT a new fact):
   - Input: \{
       "memories": [\{"memory_id": "c4184b6a", "speaker": "Alex",
         "content": "Alex is training for a marathon with a local
                     running club",
         "session_time": "9:00 am on 5 January, 2022",
         "dia_ids": ["D1:3"]\}],
       "facts": [\{"speaker": "Alex", "dia_id": "D1:9",
         "fact": "Alex is preparing for a marathon competition with
                  teammates",
         "related_memory_ids": ["c4184b6a"]\}]
     \}
   - Operations: \{"operations": []\}
   Explanation: The new fact describes the same activity already in
   "memories". It is a paraphrase, not new information -> NO
   OPERATION.

   Example (same memory shared by two facts):
   - Input: \{
       "memories": [\{"memory_id": "a32b32c1", "speaker": "John",
         "content": "John works as a software engineer",
         "session_time": "6:59 pm on 26 August, 2023",
         "dia_ids": ["D1:4"]\}],
       "facts": [
         \{"speaker": "John", "dia_id": "D3:6",
          "fact": "John changed careers to become a teacher",
          "related_memory_ids": ["a32b32c1"]\},
         \{"speaker": "John", "dia_id": "D3:7",
          "fact": "John no longer works in tech",
          "related_memory_ids": ["a32b32c1"]\}
       ]
     \}
   - Operations:
     \{"operations": [\{"operation": "UPDATE",
       "memory_id": "a32b32c1",
       "content": "John became a teacher, left software engineering",
       "dia_id": "D3:6"\}]\}
   Explanation: Both facts point to the same memory. Produce ONE
   UPDATE -- do not UPDATE the same memory_id twice.


DECISION ORDER (follow this sequence for EVERY new fact):
1. Does the new fact explicitly contradict a memory entry in its
   `related_memory_ids`? -> DELETE the contradicted entry.
2. Does a semantically equivalent entry already exist in
   `related_memory_ids` (same person, same topic, same meaning)?
   -> NO OPERATION. Stop.
3. Does an entry in `related_memory_ids` exist and the new fact
   refines, progresses, or confirms the same entity's story?
   -> UPDATE. Stop.
4. No matching entry found -> INSERT.

Follow the instruction mentioned below:
- Memory is MONOTONIC: factual information must never be lost
  unless explicitly contradicted.
- UPDATE operations MUST preserve all previously stored factual
  claims. An UPDATE must preserve all existing factual claims, but
  may rephrase them concisely within size limits.
- Do not return anything from the custom few shot prompts provided
  above.
- You should return the operations in only JSON format as shown
  above.
- Do not store small talk, greetings, generic questions. Only store
  information that conveys meaningful or significant facts.
- If there is an insert, must include speaker field. must not
  include memory_id, session_time fields because the system
  auto-generates it.
- If there is a deletion or update, must use exact memory_id from
  the "memories" list (looked up via the fact's
  `related_memory_ids`). Do not invent or guess memory IDs.
- If two facts share a related_memory_ids entry, produce at most
  ONE operation on that memory_id -- do not UPDATE or DELETE the
  same memory_id twice.
- Before outputting operations, run a strict self-check:
  1) Every `content` is understandable alone.
  2) Every `content` explicitly names the subject speaker.
  3) No unresolved vague pronouns remain.
  4) No entry is only a conversational act without durable fact
     value.

Do not return anything except the JSON format.
\end{alltt}
\end{promptbox}

\vspace{2pt}
\begin{flushleft}
{\small\textbf{Figure~\thefigure:} Prompt template for the memory
manager. The model receives the current memory store and a batch of
atomic facts (output of the fact-retrieval stage,
Appendix~\ref{app:fact-extraction-prompt}) and emits a JSON list of
\texttt{INSERT}/\texttt{UPDATE}/\texttt{DELETE} edits. A
fixed decision order, an atomicity constraint, and explicit
non-destructive update semantics together prevent the
common failure modes of LLM-based memory writers, namely fact loss,
duplicated entries, and noisy retrieval-driven overwrites.\par}
\end{flushleft}

\subsection{Prompt Template for Answer Agent (Memory Usage)}
\label{app:answer-agent-prompt-template}

The answer agent is the final stage of our pipeline. It takes a user question together with the memory entries written by the memory manager (Section~\ref{app:memory-manager-prompt}) and produces a concise, evidence-grounded answer. Two design choices are worth noting. First, each memory entry carries a timestamp, and questions in our benchmark frequently involve relative time expressions (\emph{``last year''}, \emph{``two months ago''}); the prompt therefore instructs the model to resolve such expressions to absolute dates using the timestamp of the supporting memory, rather than the question's own utterance time.
Second, the two speakers' memories are presented in separate blocks
labeled by speaker name, which prevents the model from confusing
third-party names mentioned within a memory with the speaker who owns that memory. The model is required to terminate its response with an \texttt{<answer>...</answer>} span, which is then extracted and scored against the gold answer using SQuAD-style token-level F1.
Figure~\ref{fig:prompt-template} shows the full prompt.

\refstepcounter{figure}\label{fig:prompt-template}
\begin{promptbox}[Prompt template: memory-grounded question answering]
\small\ttfamily
\begin{alltt}
You are an intelligent memory assistant tasked with retrieving accurate
information from conversation memories.
 
# CONTEXT:
You have access to memories from two speakers in a conversation. These
memories contain timestamped information that may be relevant to answering
the question.
 
# INSTRUCTIONS:
1. Carefully analyze all provided memories from both speakers
2. Pay special attention to the timestamps to determine the answer
3. If the question asks about a specific event or fact, look for direct
   evidence in the memories
4. If the memories contain contradictory information, prioritize the most
   recent memory
5. If there is a question about time references (like "last year", "two
   months ago", etc.), calculate the actual date based on the memory
   timestamp. For example, if a memory from 4 May 2022 mentions "went to
   India last year," then the trip occurred in 2021.
6. Always convert relative time references to specific dates, months, or
   years. For example, convert "last year" to "2022" or "two months ago"
   to "March 2023" based on the memory timestamp. Ignore the reference
   while answering the question.
7. Focus only on the content of the memories from both speakers. Do not
   confuse character names mentioned in memories with the actual users
   who created those memories.
8. If memories are insufficient and the question is about a general world
   fact, you may use reliable general world knowledge.
9. Keep the final answer concise, typically no more than 10-12 words; do
   not omit essential entities or dates.
 
# APPROACH (Think step by step):
1. First, examine all memories that contain information related to the
   question
2. Examine the timestamps and content of these memories carefully
3. Look for explicit mentions of dates, times, locations, or events that
   answer the question
4. If the answer requires calculation (e.g., converting relative time
   references), show your work
5. Formulate a precise, concise answer based on the evidence in the
   memories, using general world knowledge only if memories are
   insufficient
6. Double-check that your answer directly addresses the question asked
7. Ensure your final answer is specific and avoids vague time references
8. Output the final answer only in this format, with no extra text:
   <answer>YOUR_FINAL_ANSWER</answer>
 
Memories for user {{speaker_1}}:
{{speaker_1_memories}}

Memories for user {{speaker_2}}:
{{speaker_2_memories}}

Question: {{question}}
 
Answer step by step, and output the final answer in this format, with no
extra text: <answer>YOUR_FINAL_ANSWER</answer>
\end{alltt}
\end{promptbox}
 
\vspace{2pt}
\begin{flushleft}
{\small\textbf{Figure~\thefigure:} Prompt template used for memory-based question answering. Double-braced tokens denote runtime placeholders. Model outputs are parsed from the \texttt{<answer>...</answer>} span and scored with SQuAD-style token F1.\par}
\end{flushleft}

\section{Evaluation Metrics}
\label{app:evaluation_metrics}

\subsection{LLM-as-a-Judge}
\label{app:llm-judge-prompt-template}

In addition to F1, B1, we report an LLM-as-a-Judge (J) score that captures semantic equivalence between the generated answer and the gold answer, mitigating the well-known brittleness of token-level
metrics on free-form generations. We follow the judging protocol established by prior work on memory-augmented dialogue agents~\citep{mem0-chhikara2025mem0,memoryr1-yan2025memory}, and use \texttt{gpt-4o-mini} as
the judge model for all reported J scores. The judge receives the question, the gold answer, and the generated answer, and is asked to return a binary Correct/Wrong label, with explicit
instructions to be lenient toward formatting differences (e.g., \emph{``May 7''} versus \emph{``7 May''}) and toward the generated answer being more verbose than the gold. The model is required to
emit its decision as a JSON object with a single \texttt{label} field, which we parse for downstream aggregation. We chose \texttt{gpt-4o-mini} as a deliberate cost-quality trade-off: it is
strong enough to reliably handle the lenient string-matching judgments required here, while being cheap enough to run across the full evaluation set without distorting our compute budget. Figure~\ref{fig:llm-judge-prompt} shows the full prompt.

\refstepcounter{figure}\label{fig:llm-judge-prompt}
\begin{promptbox}[Prompt template: LLM-as-a-Judge]
\small\ttfamily
\begin{alltt}
Your task is to label an answer to a question as 'CORRECT' or 'WRONG'. 
You will be given:
(1) a question, (2) a gold (ground truth) answer, (3) a generated answer.

The gold answer is usually concise; the generated answer may be longer. 
Be generous: if the generated answer touches on the same topic/date as 
the gold, count CORRECT. Different formats for the same date (e.g. "May 7" 
vs "7 May") are CORRECT.

Question: \{question\}
Gold answer: \{gold_answer\}
Generated answer: \{generated_answer\}

First give a one-sentence reasoning, then finish with CORRECT or WRONG. Do 
NOT include both.

Return a JSON object with key "label" whose value is exactly "CORRECT" 
or "WRONG".
\end{alltt}
\end{promptbox}

\vspace{2pt}
\begin{flushleft}
{\small\textbf{Figure~\thefigure:} Prompt template for the LLM-as-a-Judge evaluator, instantiated with \texttt{gpt-4o-mini}. Single-braced tokens (\texttt{\{question\}}, \texttt{\{gold\_answer\}},
\texttt{\{generated\_answer\}}) are runtime placeholders.\par}
\end{flushleft}

\subsection{Memory-Failure Rate (M-Fail)}
To diagnose memory-bank construction quality beyond answer-level metrics, we define \textbf{M-Fail} as the fraction of gold evidence that is missing from the memory bank. For each question $q$, let $\mathcal{E}_q \subseteq \mathcal{D}$ be the set of gold evidence dialogue-turn IDs required to answer $q$, and let $\mathcal{M} \subseteq \mathcal{D}$ denote the set of dialogue-turn IDs currently stored in the memory bank. We compute
\begin{equation}
\mathrm{M\text{-}Fail}
=
\frac{\sum_q \left| \mathcal{E}_q \setminus \mathcal{M} \right|}
{\sum_q |\mathcal{E}_q|}.
\end{equation}
Lower is better; $\mathrm{M\text{-}Fail}=0$ means all required evidence is present in memory. This metric isolates \emph{memory-construction} errors (evidence never stored) from downstream retrieval or answer-generation effects.

\section{Limitations and Future Work}
\label{app:limitations_future_work}
Our study focuses on long-horizon text-only multi-session dialogue, where memory is constructed from conversational turns and evaluated through downstream QA. Extending the same training paradigm to multimodal settings, such as image-grounded dialogue, video memory, or embodied interaction, remains unexplored. While the proposed credit-assignment framework is in principle agnostic to the modality of memory content, its practical effectiveness in richer environments still requires further investigation.

This work has both potential positive and negative societal impacts. On the positive side, Memory-R2 improves the training of long-horizon memory-augmented LLM agents, which may help make current memory systems more reliable, consistent, and practically useful. More effective memory construction can benefit applications such as personalized assistants, long-term educational support, and tools that require maintaining user context across multiple interactions. More broadly, improving memory quality may reduce failure modes caused by forgetting or inconsistent recall, which are common limitations of current LLM-based agents.

At the same time, stronger persistent memory systems also raise important risks. In real-world deployments, memory mechanisms may store sensitive personal information over extended interactions, creating privacy and security concerns if such information is retained unnecessarily, retrieved inappropriately, or exposed to unauthorized parties. As memory becomes more accurate and durable, these risks may become more consequential. In addition, errors in stored memory may persist across sessions and affect later interactions. We therefore view privacy-preserving memory design, secure storage and access control, and safer memory-update mechanisms as important directions for future work.
\section{Additional Ablations and Generalization Results}

This section provides additional evidence for the effectiveness and stability of Memory-R2. 

\begin{figure}[t]
  \centering
  \includegraphics[width=1\textwidth]{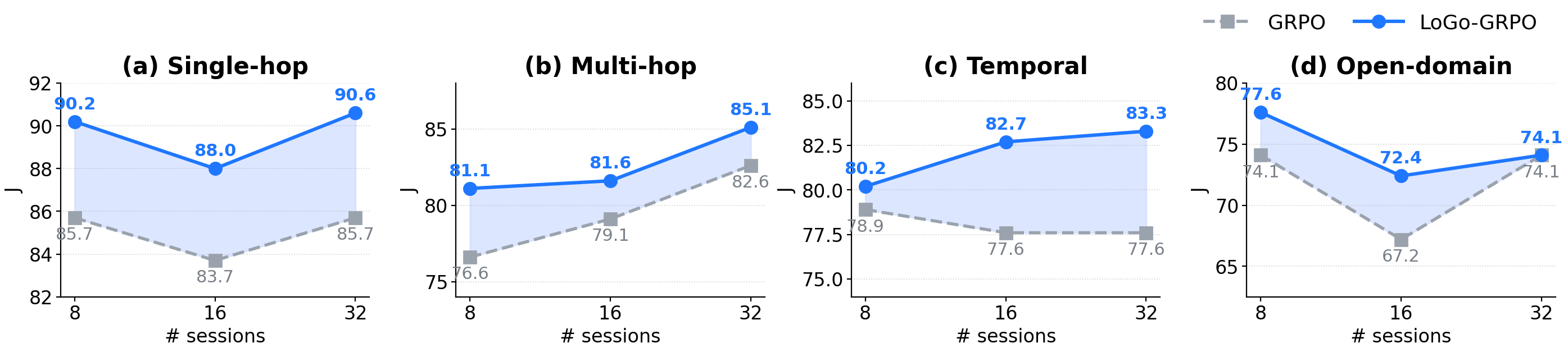}
  \caption{\textbf{LoGo-GRPO consistently outperforms GRPO across all question types and curriculum stages.} 
  Judge accuracy (\textbf{J}) at curriculum stages 8\,$\rightarrow$\,16\,$\rightarrow$\,32 sessions, broken down by question type: \textbf{(a)} Single-hop, \textbf{(b)} Multi-hop, \textbf{(c)} Temporal, \textbf{(d)} Open-domain. LoGo-GRPO (blue) dominates GRPO (gray) at every stage and on every category, with the shaded band visualizing the gap, indicating that local rerollouts constantly mitigate credit-assignment bias. Accuracy continues to improve with the session range on long-horizon question types (Multi-hop, Temporal), while remaining stable on the others.}
  \label{fig:logo_grpo_per_type}
\end{figure}

Figure~\ref{fig:logo_grpo_per_type} breaks down LoGo-GRPO versus standard GRPO across question types and curriculum stages, showing consistent gains from local rerollouts. 

\begin{figure}[t]
  \centering
  \includegraphics[width=1\textwidth]{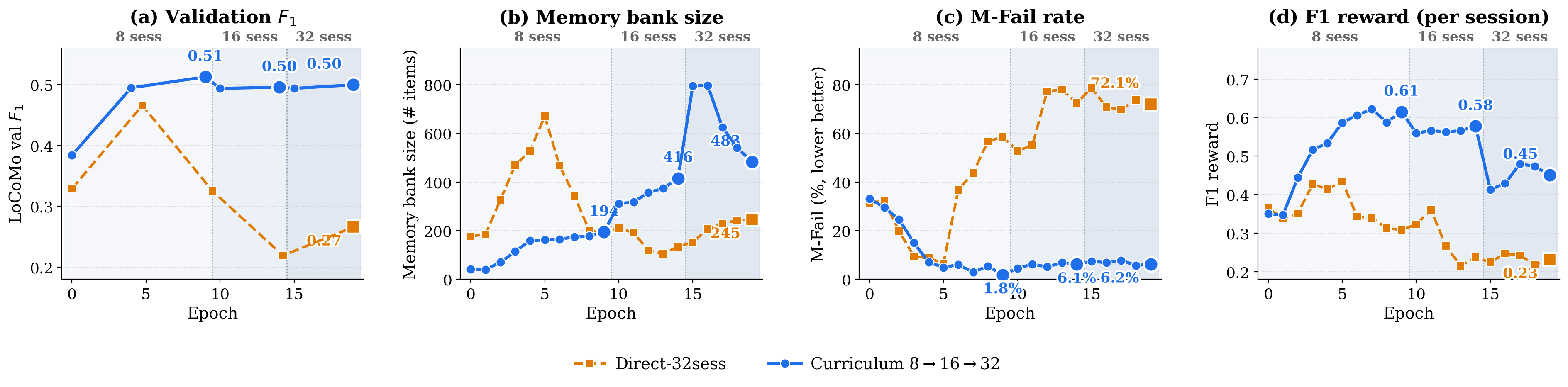}
  \caption{\textbf{Curriculum learning is essential for stable long-horizon training.} 
  Training dynamics of curriculum 8\,$\rightarrow$\,16\,$\rightarrow$\,32 sessions (blue) vs.\ direct 32-session training (orange) under equal compute. The x-axis is the cumulative epochs within the curriculum; direct-32sess is linearly stretched onto the same axis for fair comparison.
  \textbf{(a)} Validation $F_1$ on LoCoMo: the curriculum stabilizes around $0.50$, while direct-32sess collapses from a peak of $0.47$ down to $0.27$.
  \textbf{(b)} Memory bank size grows steadily under the curriculum ($194\,\rightarrow\,416\,\rightarrow\,483$ items at stage ends), reflecting healthy accumulation; the direct run instead inflates and then truncates erratically.
  \textbf{(c)} Memory failure rate (M-Fail) stays below $7\%$ throughout the curriculum, but explodes to over $70\%$ for direct-32sess once training enters the long-horizon regime.
  \textbf{(d)} Per-session $F_1$ reward remains high across the curriculum ($0.61 / 0.58 / 0.45$ at stage ends), while the direct run degrades to $0.23$.
  Together, these dynamics show that early errors in the long-horizon setting propagate across sessions and corrupt the memory bank, and that the curriculum allows the policy to acquire reliable short-horizon memory behaviors before tackling longer trajectories.}
  \label{fig:direct_vs_curriculum}
\end{figure}

Figure~\ref{fig:direct_vs_curriculum} compares the proposed 8$\rightarrow$16$\rightarrow$32-session curriculum with direct 32-session training under equal compute, demonstrating that curriculum learning stabilizes validation performance, memory growth, memory quality, and reward signals. 

\begin{figure}[t]
    \centering
    \includegraphics[width=\textwidth]{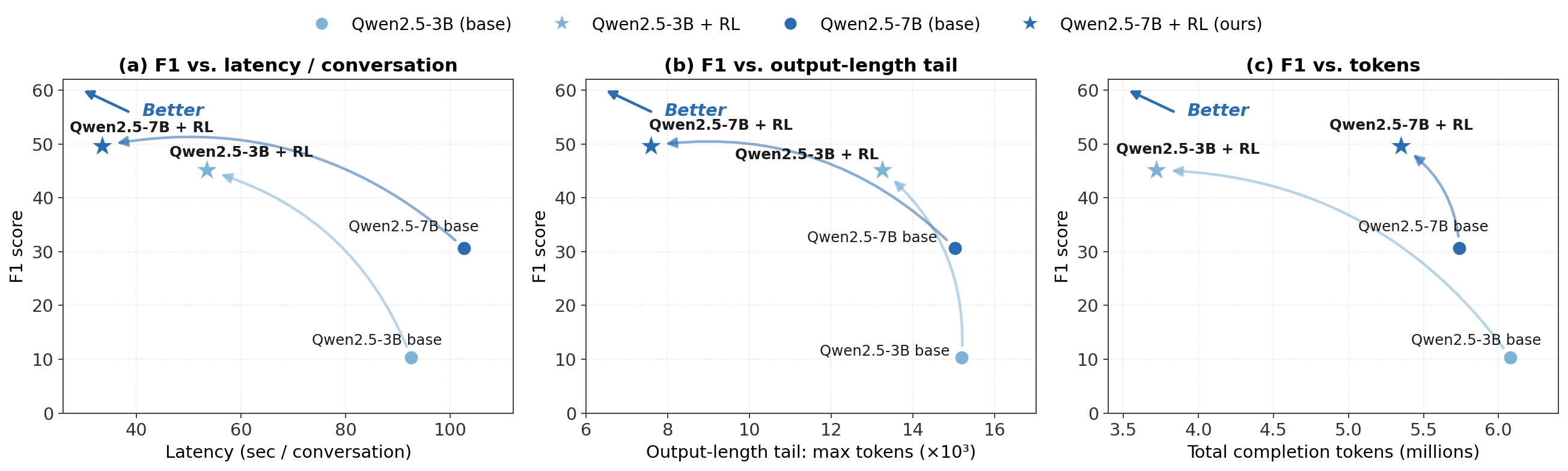}
    \caption{Latency-mechanism diagnostics. (a) Memory-R2 improves F1 while reducing per-conversation latency. (b) Memory-R2 shortens the output-length tail, indicating fewer overly long generations. (c) Memory-R2 reduces total completion tokens, especially for Qwen2.5-3B, indicating a more concise memory-construction policy. Together, these diagnostics suggest that the latency gains come from more controlled and less redundant generation.}
    \label{fig:latency_mechanism}
\end{figure}

Figure~\ref{fig:latency_mechanism} further diagnoses the latency improvements of Memory-R2, showing that the 3B model reduces latency mainly through more concise generations, while the 7B model reduces latency by suppressing long output-length tails and improving batched decoding efficiency.

Table~\ref{tab:main_result_variance} reports mean and standard deviation over three independent runs, confirming that the improvements of Memory-R2 are robust across random seeds.

\label{app:additonal_ablations}

\begin{table}[t]
\centering
\small
\setlength{\tabcolsep}{4pt}
\caption{Main results per question category, averaged over 3
independent runs with different random seeds. We report mean $\pm$
standard deviation. The J score is computed by an LLM-as-a-Judge with
\texttt{gpt-4o-mini} as the judge model
(Section~\ref{app:llm-judge-prompt-template}).}
\label{tab:main_result_variance}
\begin{tabular}{llccc}
\toprule
\textbf{Model} & \textbf{Category} & \textbf{F1} $\uparrow$ & \textbf{B1} $\uparrow$ & \textbf{J} $\uparrow$ \\
\midrule
\multirow{5}{*}{Memory-R2}
  & Single-hop    & 54.06 $\pm$ 2.12 & 48.73 $\pm$ 2.26 & 86.80 $\pm$ 0.48 \\          
  & Multi-hop     & 38.41 $\pm$ 0.91 & 30.90 $\pm$ 0.46 & 80.93 $\pm$ 1.52 \\          
  & Temporal      & 59.65 $\pm$ 1.00 & 50.05 $\pm$ 1.10 & 69.90 $\pm$ 1.34 \\          
  & Open-domain   & 20.76 $\pm$ 1.06 & 16.78 $\pm$ 0.66 & 67.53 $\pm$ 2.04 \\          
  & Overall       & 50.60 $\pm$ 1.34 & 44.01 $\pm$ 1.37 & 80.99 $\pm$ 0.28 \\               
\midrule
\multirow{5}{*}{Memory-R2 (OSS)}
  & Single-hop    & 50.98 $\pm$ 0.45 & 45.79 $\pm$ 0.48 & 90.58 $\pm$ 1.56 \\
  & Multi-hop     & 36.37 $\pm$ 1.63 & 29.79 $\pm$ 1.91 & 85.07 $\pm$ 1.41 \\
  & Temporal      & 62.48 $\pm$ 0.52 & 55.33 $\pm$ 0.42 & 83.33 $\pm$ 1.49 \\
  & Open-domain   & 30.22 $\pm$ 0.71 & 24.50 $\pm$ 1.11 & 74.14 $\pm$ 0.00 \\
  & Overall       & 49.67 $\pm$ 0.60 & 43.77 $\pm$ 0.59 & 87.10 $\pm$ 1.43 \\
\bottomrule
\end{tabular}
\end{table}

\clearpage


\end{document}